\documentclass[11pt]{article}
\usepackage{aupreprint}

\title{Agentic University: Collaborative Agents for Personalized Learning}

\newcommand{\AUAuthors}{%
  {\fontsize{11}{15}\selectfont
  Jingru Fan\textsuperscript{1}\quad Tianle Zhou\textsuperscript{2}\quad
  Yunjie Zhou\textsuperscript{1}\quad Zihao He\textsuperscript{1}\quad
  Yixuan Zhou\textsuperscript{1}\\[2pt]
  Kezhao Wu\textsuperscript{1}\quad Xinyu Zheng\textsuperscript{1}\quad
  Xiyuan Wang\textsuperscript{3}\quad Yuheng Wang\textsuperscript{1}\quad
  Zhi Li\textsuperscript{1}\\[2pt]
  Songhua Liu\textsuperscript{1}\quad Weinan E\textsuperscript{1}\quad
  Chen Qian\textsuperscript{1}\par}
  \vspace{8pt}
  {\fontsize{9.5}{13}\selectfont
  \textsuperscript{1}Shanghai Jiao Tong University\qquad
  \textsuperscript{2}Nanyang Technological University\\
  \textsuperscript{3}National University of Singapore\par}
  \vspace{5pt}
  {\fontsize{9}{12}\selectfont
  \href{mailto:fanjingru510@sjtu.edu.cn}{fanjingru510@sjtu.edu.cn}\qquad
  \href{mailto:qianc@sjtu.edu.cn}{qianc@sjtu.edu.cn}\par}
}
\author{\AUAuthors}
\date{}
\hypersetup{
  pdftitle={Agentic University: Collaborative Agents for Personalized Learning},
  pdfauthor={Jingru Fan, Tianle Zhou, Yunjie Zhou, Zihao He, Yixuan Zhou, Kezhao Wu, Xinyu Zheng, Xiyuan Wang, Yuheng Wang, Zhi Li, Songhua Liu, Weinan E, Chen Qian}
}

\begin{document}
\maketitle
\begin{abstract}
Large language models support personalized tutoring, content generation,
and adaptive learning paths. However, capability development requires
these resources and interactions to form a coherent learning process
that evolves with the learner. A central challenge is to align goals,
activities, and assessment while using learning evidence to decide what
should follow. We introduce Human-Participatory Multi-Agent Organization
and instantiate it in Agentic University (AU). In AU, learners participate
as organization members and work with specialized agents to maintain a
revisable Capability Program. The program connects personal goals,
learning activities across modalities, and layered assessment, while
learner feedback and performance guide subsequent support. Our evaluation
shows advantages in program organization, assessment design, and revision
quality. Users achieved larger mean assessment gains (12.19 versus 5.61
points on a 50-point scale) and reported stronger cognitive support. This
work contributes an organizational approach to adaptive learning that
makes learner participation a continuing basis for multi-agent
coordination.\par\smallskip\noindent The Agentic University project is available at \href{https://github.com/AgenticOrgLab/AgenticUniversity.git}{\raisebox{-0.45em}{\includegraphics[height=1.8em]{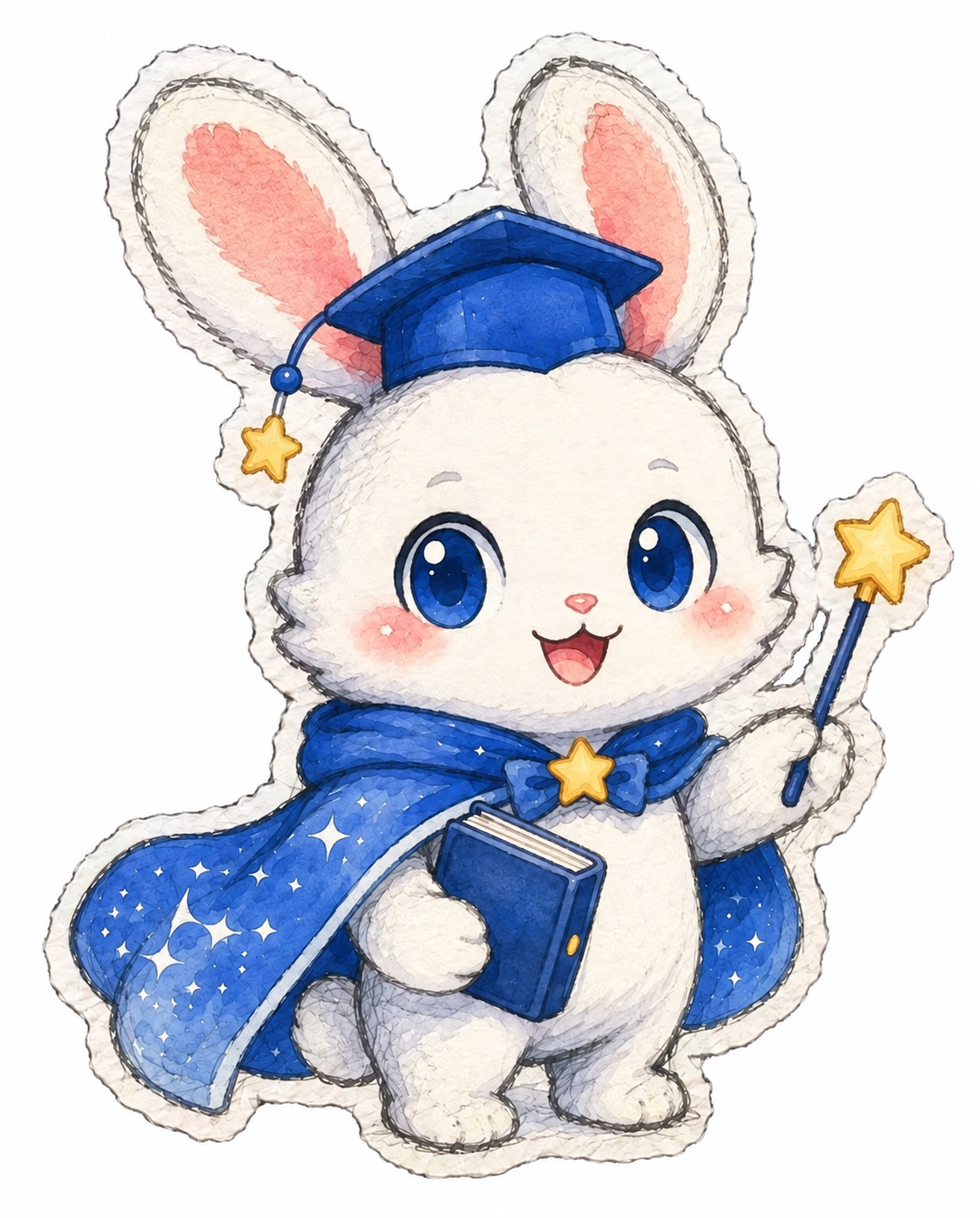}}}\hspace{0.4em}\url{https://github.com/AgenticOrgLab/AgenticUniversity.git}. \end{abstract}

\section{Introduction}
\label{sec:introduction}

Advances in large language models (LLMs) enable educational systems to
provide personalized tutoring, generate instructional materials, and plan
learning activities~\cite{jurenka2024responsible,kestin2025aitutoring,wang2025genmentor}.
Recent applications support instructional design~\cite{zhang2025eduplanner},
teaching video generation~\cite{wang2026teachmaster}, and real-time guidance
for human tutors~\cite{wang2024tutorcopilot}.
Other systems use learner profiles and feedback to plan and refine
learning paths~\cite{wang2025genmentor,xu2026malpp}, and use interaction
history to support ongoing tutoring and practice~\cite{zhao2026deeptutor}.

However, for learners seeking to develop their capabilities, generating
instructional materials or conducting tutoring dialogues provides only
part of the required support.
These materials and interactions need to form a coherent learning process
that aligns goals, learning activities, and assessment~\cite{biggs1996enhancing}.
A system therefore needs to coordinate how different learning artifacts
support an individual learner.
As learning evidence accumulates, it needs to adjust the content of
subsequent activities and how they connect.
A learning path organizes tasks, and instructional content and practice
support those tasks.
Evidence-centered assessment connects claims about learner capability to
observable evidence and the tasks that elicit it~\cite{mislevy2003brief}.
Recent work applies this approach to evidence from human-AI writing
interactions~\cite{cheng2024evidence}.
Assistance conditions also matter.
A study of high-school mathematics found that better performance during
AI-assisted practice did not necessarily translate into better unaided
performance~\cite{bastani2025guardrails}.
Difficulties identified during learning should guide subsequent support,
and assessed performance should inform revisions to the learning
path~\cite{black1998assessment}.
We therefore examine how to organize sustained learning around each
learner's goals, prior knowledge, and progress.

\begin{figure}[t]
    \centering
    \includegraphics[height=6cm]{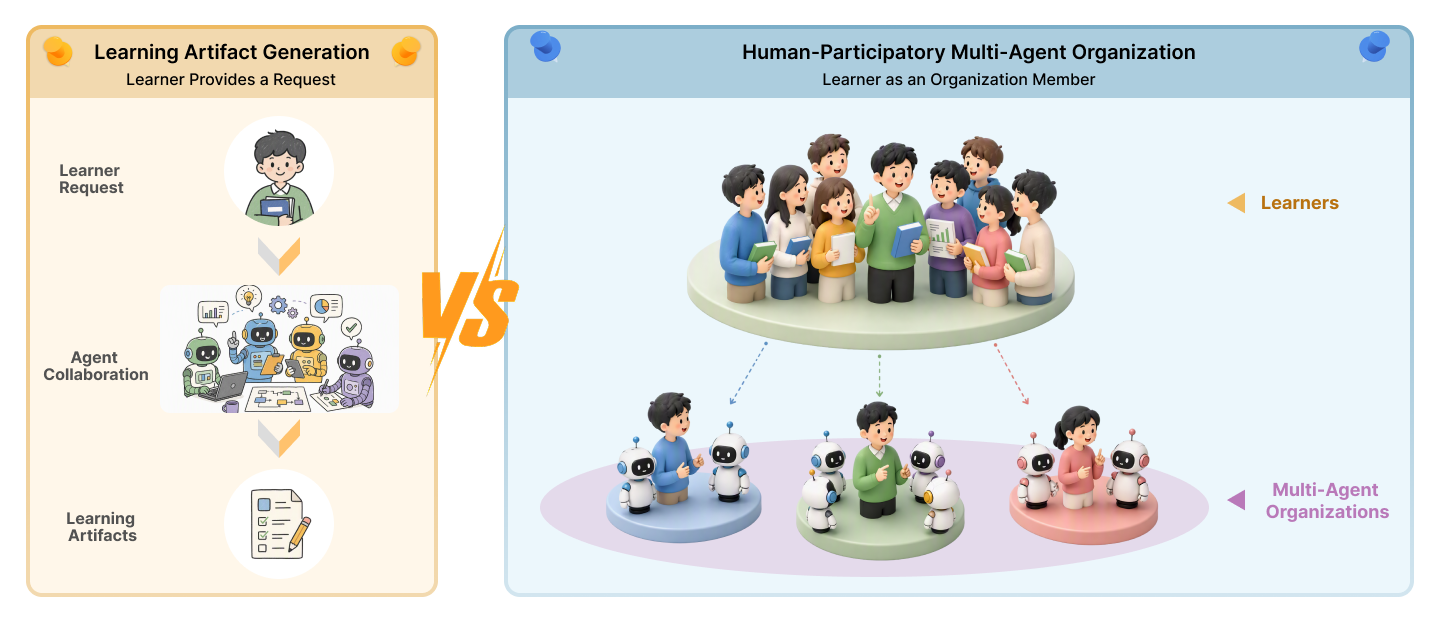}
    \caption{Two ways of organizing multi-agent work around a learner.
    Left: a learner provides a request, and agents collaborate to generate
    learning artifacts. Right: a learner participates as a member of a
    multi-agent organization. Its activities support learning, while
    learner feedback and performance inform subsequent work.}
    \Description{The left panel shows a learner providing a request,
    a group of agents collaborating, and the resulting learning artifacts.
    The right panel shows a group of learners above three smaller groups,
    each containing one learner and several agents. Dotted arrows connect
    the learners above to these groups.}
    \label{fig:organization-shift}
\end{figure}

We build on role specialization and collaboration in LLM-based multi-agent
systems (MAS)~\cite{chen2024agentverse,wang2025genmentor} to introduce Human-Participatory Multi-Agent Organization.
The framework draws on artifacts as a means of coordinating cooperative
work~\cite{schmidt1996mechanisms}.
Its emphasis on learner participation is informed by research on learner
decision making~\cite{vincoli2025agency} and human-AI interaction guidelines
on feedback and control~\cite{amershi2019guidelines}.
We use this framework to develop the Agentic University (AU) system.
AU is designed for learners pursuing specific capability goals, with a
focus on university students.
As Figure~\ref{fig:organization-shift} shows, each learner participates as
an organization member and jointly maintains a revisable Capability Program with agents.
The program connects personal goals, diverse learning activities across
modalities, and assessment requirements.
The organization provides learning support through a learning path, an
interactive classroom, and layered assessment, and adapts subsequent
learning activities.

We evaluated AU through three studies of program formation,
evidence-guided revision, and human use.
Model-based evaluation identified program organization and assessment
design as AU's clearest advantages.
In revision comparisons, AU received higher ratings for revision
appropriateness, program improvement, and path coherence.
Both systems received high ratings for alignment between evidence and
revisions.
In the human study, we observed mean assessment gains of 12.19 points for
AU users and 5.61 points for the comparison group on a 50-point scale.
AU users also reported stronger cognitive support.

This paper makes three contributions:
\begin{enumerate}
    \item We introduce Human-Participatory Multi-Agent Organization,
    a framework in which learners participate as organization members
    and work with agents to organize and adapt ongoing learning activities.

    \item We develop AU, which uses the Capability Program to connect
    personal goals, diverse learning activities across modalities, and
    layered assessment to provide adaptive learning support.

    \item We report three studies of program formation, evidence-guided
    revision, and human use, providing empirical evidence on program
    quality, revision quality, and learner outcomes and experiences.
\end{enumerate}

\section{Related Work}
\label{sec:related-work}

\subsection{From Personalized Tutoring to Coherent Learning Programs}

Personalized learning research connected representations of the learner to instructional decisions.
Knowledge tracing estimated skill acquisition from performance~\cite{corbett1995knowledgetracing}, and deep knowledge tracing learned representations of learning histories with recurrent neural networks~\cite{piech2015deep}.
Adaptive hypermedia used learner models to adapt content presentation and navigation~\cite{brusilovsky2001adaptive}, while AutoTutor used mixed-initiative dialogue to support learning~\cite{graesser2005autotutor}.
Together, these approaches established personalization at the levels of learner state, content access, and instructional interaction.

Recent systems extended personalization through pedagogically informed generation and course-specific support.
LearnLM incorporated learning-science principles into model development and pedagogical evaluation~\cite{jurenka2024responsible}, while CourseAssist aligned tutoring responses with course materials and learning objectives through retrieval and question decomposition~\cite{feng2024courseassist}.
Other systems broadened the available learning activities through customized instructional design~\cite{zhang2025eduplanner}, teaching-video production~\cite{wang2026teachmaster}, and agent-supported online classrooms~\cite{yu2026maic}.
Together, these systems expanded the instructional resources and interactions available for personalized learning.

A further line coordinated personalization across learning paths.
MALPP coordinated learner analytics, path planning, and reflection agents to iteratively refine personalized paths with explicit rationales~\cite{xu2026malpp}, while Pxplore combined goal-driven learner-state modeling with policy training for personalized path planning~\cite{lim2026plpp}.
GenMentor, our closest system-level comparison, connected goal-to-skill mapping, dynamic learner profiles, instructional content, and quizzes.
It revised learning paths using updated profiles and required learner approval before applying changes~\cite{wang2025genmentor}.
DeepTutor shared evolving learner profiles and interaction memory across tutoring functions~\cite{zhao2026deeptutor}.
These systems established precedents for adaptive paths, continuing personalization, and coordinated support.
AU extended this line through a Human-Participatory Multi-Agent Organization for capability development.
The learner participated as an organization member alongside agents responsible for planning, mentoring, assessment, and replanning.
Together, they maintained a revisable Capability Program that connected personal goals, multimodal learning activities, and layered assessment.
The organization used learner contributions and performance evidence to adapt subsequent learning, with the learner retaining control over path changes.

\subsection{From Role Specialization to Shared-State Coordination in Multi-Agent Systems}

Research on multi-agent organizations formalized collective activity through agents, groups, and roles~\cite{ferber2003organizations}.
MOISE+ further distinguished structural, functional, and deontic specifications, connecting organizational roles to collective goals and obligations~\cite{hubner2002moise}.
LLM-based systems operationalized complementary forms of collaboration.
CAMEL used role-playing to structure agent interaction~\cite{li2023camel}, and AgentVerse coordinated teams of agents~\cite{chen2024agentverse}.
MetaGPT organized intermediate artifacts through standard operating procedures~\cite{hong2023metagpt}, while ChatDev linked specialized roles through a staged software-development process~\cite{qian2024chatdev}.
These approaches provided precedents for organizing specialized outputs into collective work.

Later work examined how teams coordinated during execution.
AutoGen supported conversations among agents, tools, and people~\cite{wu2024autogen}.
Internet of Agents supported heterogeneous agents through dynamic teaming~\cite{chen2025internetagents}, and Optima trained agents to balance communication efficiency and task effectiveness~\cite{chen2025optima}.
Magentic-One maintained task and progress ledgers to guide execution and replanning~\cite{fourney2024magenticone}.
Cemri et al.\ identified failures involving system design, inter-agent misalignment, and verification in multi-agent execution traces~\cite{cemri2025masfail}.
Together, these studies highlighted the importance of coordination rules and shared context alongside role specialization.

Shared artifacts provided a further basis for coordination.
Schmidt and Simone described coordination mechanisms as artifacts coupled with conventions for managing interdependent work~\cite{schmidt1996mechanisms}.
Recent work examined the authority and consistency of persistent state: governed collaborative memory distinguished private from institutional memory and connected memory admission to ratification, provenance, and version lineage~\cite{cuadros2026governed}; the Artifact Coherence System serialized writes to each artifact and used versioning and invalidation to coordinate cached copies~\cite{parakhin2026tokencoherence}.
AU applied an organizational perspective to educational dependencies: its Capability Program connected objectives, prerequisites, activities, assessment requirements, and current program state.
Its evidence records retained sources, conditions, and permitted updates, while its revision history linked evidence and diagnosis to learner-confirmed changes and path versions.

\subsection{Human Participation and Learner Agency}

Research on learner agency examined how learners influenced AI-supported activity.
Vincoli et al.\ distinguished agency over content, feedback and help, co-orchestration, and data~\cite{vincoli2025agency}.
Dai and Lai identified initiating and redirecting, mindful adoption, external help-seeking, and reflection as interrelated aspects of agency in students' AI-assisted learning~\cite{dai2025agency}.
These perspectives situated participation across learning decisions and interactions.
Open learner modeling offered a concrete mechanism by making learner representations inspectable and, in some designs, negotiable or editable~\cite{bull2016smili}.

Educational systems also distributed participation across different human roles.
AI2T supported tutor authoring through demonstrations and corrective feedback~\cite{weitekamp2024ai2t}, while TeachTune helped teachers inspect pedagogical agents across simulated learner profiles~\cite{jin2025teachtune}.
During instruction, Tutor CoPilot supplied suggestions to human tutors~\cite{wang2024tutorcopilot}, and the LearnLM classroom trial placed every generated message under expert review before delivery~\cite{learnlm2025classroomrct}.
These systems located human judgment in tutor construction, review, and instructional delivery.

Human--AI interaction research addressed related questions of initiative and control.
Mixed-initiative interaction examined how people and systems shared initiative~\cite{horvitz1999principles}, and human--AI interaction guidelines emphasized correction, feedback, and control over time~\cite{amershi2019guidelines}.
AgentCoord enabled users to inspect and revise multi-agent plans, agent assignments, and task processes~\cite{pan2025agentcoord}.
In education, LearnMate enabled learners to adjust personalized plans and schedules while receiving contextual support~\cite{wang2025learnmate}.
These systems demonstrated concrete ways for people to shape plans and agent behavior.

AU investigated how learner contributions remained consequential when educational work passed between roles.
Its Human Node retained goals, constraints, and corrections, while its Capability Program connected those contributions to subsequent planning, mentoring, assessment, and replanning.
Learner confirmation governed the application of path changes.
This design treated continuing participation as part of organizational coordination.

\subsection{Connecting Capability Assessment to Program Revision}

Educational research provided principles for connecting goals, instructional activity, and evidence.
Constructive alignment linked intended outcomes, learning activities, and assessment~\cite{biggs1996enhancing}.
Evidence-centered design connected capability claims to observable evidence and the tasks that elicited it~\cite{mislevy2003brief}, while authentic-assessment research situated assessment in meaningful tasks, contexts, and criteria~\cite{gulikers2004framework}.
These perspectives informed AU's alignment of knowledge-point checks, stage micro-projects, and an integrative simulation with objectives at corresponding scopes.

AI assistance made the interpretation of performance especially consequential.
Cheng et al.\ used evidence-centered assessment to examine human--AI writing processes~\cite{cheng2024evidence}.
Bastani et al.\ found that better performance during AI-assisted mathematics practice did not necessarily imply better performance after assistance was removed~\cite{bastani2025guardrails}.
Formative-assessment research connected evidence to subsequent teaching and learning decisions~\cite{black1998assessment,black2009developing}.
Sadler further emphasized learners' understanding of quality, comparison of their work against that standard, and action to close the gap~\cite{sadler1989formative}.
These perspectives motivated AU's attention to what an observation supported and how it informed subsequent instruction.
AU distinguished Learning-State Evidence for interpreting support needs from Capability Evidence for updating estimates within the assessed task and assistance conditions.
Both informed targeted program revisions, with explicit links to the affected objectives, activities, and assessment requirements.

\section{Formative Study}
\label{sec:formative-study}

Learning support needs to connect learners' backgrounds and goals
with their activities, progress, and next steps. We conducted
formative interviews to understand how people make these connections
in study, technical work, and teaching. The findings informed AU's
design goals for maintaining learner context, organizing learning
activities, interpreting performance, and adapting subsequent work.

\subsection{Participants and Procedure}
\label{sec:formative-participants}

\paragraph{Recruitment and sampling rationale.}
We recruited seven participants through online social platforms.
We used purposive sampling~\cite{palinkas2015purposeful} to select
people with relevant experience in learning, applying, or teaching
technical knowledge. Their experience could inform the study of learning planning,
progress judgments, and responses to difficulty.

Five primary learners (L1 to L5) provided accounts of managing their
own learning. Their backgrounds included undergraduate computing,
AI exploration and career preparation, extensive online-course
learning, and interdisciplinary doctoral study.
We included a technical practitioner (L6) with data-pipeline experience
to examine learning shaped by workplace delivery needs.
We included a data structures instructor (E1) to examine how educators
interpret learner performance and select instructional support.
These backgrounds provided variation in learning goals, experience,
and settings. We distinguished L6's workplace account and E1's
instructional observations from the primary learners' accounts
throughout the analysis.

\paragraph{Interview procedure and ethics.}
Individual online interviews took place on February 25 and 26, 2026,
in Mandarin Chinese.
Each interview lasted 40 to 60 minutes.
We used a common semi-structured guide to ask how participants
planned learning, judged progress, and responded to difficulty.
Follow-up questions explored these practices in each participant's
learning, workplace, or teaching context. Participants discussed
their existing practices without viewing a prototype.

The university's ethics committee approved the study.
Researchers explained the study purpose, procedure, recording
arrangements, intended data use, and participant rights.
All participants provided electronic informed consent before
their interviews and agreed to audio recording.
Each participant received CNY~100 through online payment.
Researchers checked the automatically generated transcripts against
the recordings and corrected transcription errors.
We removed or generalized identifying details and replaced
names with participant identifiers before analysis.

\subsection{Analysis}
\label{sec:formative-analysis}

Two researchers manually analyzed the transcripts using inductive
thematic analysis at the semantic level~\cite{braun2006using}.
They open-coded passages to capture participants' explicit accounts
and developed a shared codebook through repeated coding, review,
and revision~\cite{macqueen1998codebook}. A passage could receive
multiple codes. The researchers compared interpretations, revisited
the surrounding transcript, and discussed differences to refine
the codes and their application.

The researchers grouped the codes into themes and reviewed supporting
and contrasting accounts. They kept direct experiences, prospective
accounts, and instructor observations distinct when interpreting
each theme. The final analysis comprised 22 codes and four themes.
We retained the coding records, analytic decisions, and supporting
excerpts as an audit trail~\cite{nowell2017thematic}.

\subsection{Findings}
\label{sec:formative-findings}

Italic quotations are English translations of Mandarin interview excerpts.

\subsubsection{F1: Learner context was fragmented and sometimes inaccurate}

Participants carried context across courses, AI conversations, and other resources. L1 explained,
\emph{``In a new conversation, I usually have to explain my background
again, how far I have learned, and what I still do not understand.''}
L3 reported that tools lacked context from learning elsewhere. E1 noted that scores and submissions did not convey students' specific difficulties between instructors.

L5's humanities label also led AI-generated routes to discount prior STEM training, while recommendations for L2 assumed continued commitment to a briefly explored direction. These accounts call for context continuity and correction of inaccurate or outdated assumptions.

\subsubsection{F2: Learning routes left gaps in action and completion criteria}

L2 wanted a concrete daily task within a short planning horizon; L3 found generated routes fragmented and poorly connected to resources. L1 explained,
\emph{``The route looks clear, but when I execute it, it is difficult
to judge what counts as finished.''}
L5 reprioritized learning amid competing doctoral responsibilities. L6 planned workplace learning around immediate delivery demands, while L4 valued fixed courses linked to examination or work goals and including practice. These accounts call for linking broad goals to feasible activities and explicit completion criteria, with planning horizons suited to learners' circumstances.

\subsubsection{F3: Capability judgments depended on tasks and available support}

Participants distinguished activity completion from applying knowledge. L1 could follow a model-training video but struggled with missing values, feature selection, and evaluation on an actual dataset. E1 used explanations and open-ended tasks to assess understanding beyond participation, submissions, or running code, revealing further difficulties.

Available assistance also shaped performance judgments. L3 explained,
\emph{``If I follow the quick start, the project will usually run,
and AI helps me repair pipeline problems. So the project running
by itself does not say much.''}
L6 described a different workplace criterion:
\emph{``If you can bring the resources together and complete the
work as the owner, that is also a form of workplace capability.''}
These accounts distinguish independent technical execution from responsibility for completing work with available resources. Performance evidence therefore needs to be interpreted in relation to the task and support used. L5 also described possible causes of non-completion: knowledge gaps, limited time, competing priorities, and unsuitable resources.

\subsubsection{F4: Interpreting difficulties guided subsequent learning choices}

L2 described how difficulties guided subsequent learning:
\emph{``I treat it as a new judgment about the boundary of my capability,
and then decide whether to strengthen it or adjust what I learn next.''}
L3 gave AI questions and self-produced answers to identify weaknesses and generate follow-up questions. L4 changed explanations or resources when a course did not resolve a difficulty.

E1 distinguished conceptual misunderstanding from insufficient practice when selecting support and used conversation to interpret conflicting assessment results. These interpretations guided further explanation, practice, resource changes, or revised priorities.

\subsection{Design Goals}
\label{sec:formative-design-goals}

The four themes and related work informed AU's design goals and corresponding system support.

\paragraph{DG1: Maintain a persistent and correctable learner representation (F1).}
AU should retain relevant learner context across interactions and roles, record its sources, and allow learners to correct inaccurate or outdated assumptions. This goal informed the Human Node and selective SharedMemory.

\paragraph{DG2: Connect goals to activities, completion criteria, and progress (F2).}
AU should connect goals to activities, resources, prerequisites, and explicit completion criteria, consistent with constructive alignment among intended learning outcomes, learning activities, and assessment~\cite{biggs1996enhancing}. It should track completed, remaining, and deferred work within a program scope suited to the learner's planning horizon. This goal informed the Capability Program.

\paragraph{DG3: Match capability claims to the scope of evidence (F3).}
AU should interpret performance in relation to the target capability, task, conditions, timing, and allowed support. This goal informed its assessment design and the distinction between Learning-State Evidence and Capability Evidence.

\paragraph{DG4: Use learner evidence and input to guide program revision (F4).}
AU should use new evidence and learner input to revise subsequent work, following formative assessment's use of learning evidence to guide the next instructional step~\cite{black2009developing}. Revisions should address identified difficulties, preserve relevant work, and allow learner inspection. AU should support learner review when evidence conflicts or time and other constraints may explain performance.

\section{Agentic University}
\label{sec:tu-integrated}
\label{sec:tu-interaction}

Agentic University (AU) supports learners from goal setting through
learning and program revision.
A learner arrives with a goal and begins in a guided conversation,
describing background, preferences, and constraints, and correcting the
system's assumptions along the way. AU organizes a learning program of
stages and knowledge-point nodes. Within each node, the learner studies
materials, watches instructional videos, and answers Pause-QA questions,
turning to mentors for explanations, examples, or hints as difficulties
arise. Node assessments, stage micro-projects, and a final enterprise
simulation follow, each returning feedback the learner can act on. When assessment results or learning interactions indicate a need for
additional support, AU considers a revision to the remaining path.
It applies proposed changes after learner confirmation.

\begin{figure*}[t]
  \centering
  \includegraphics[width=\textwidth]{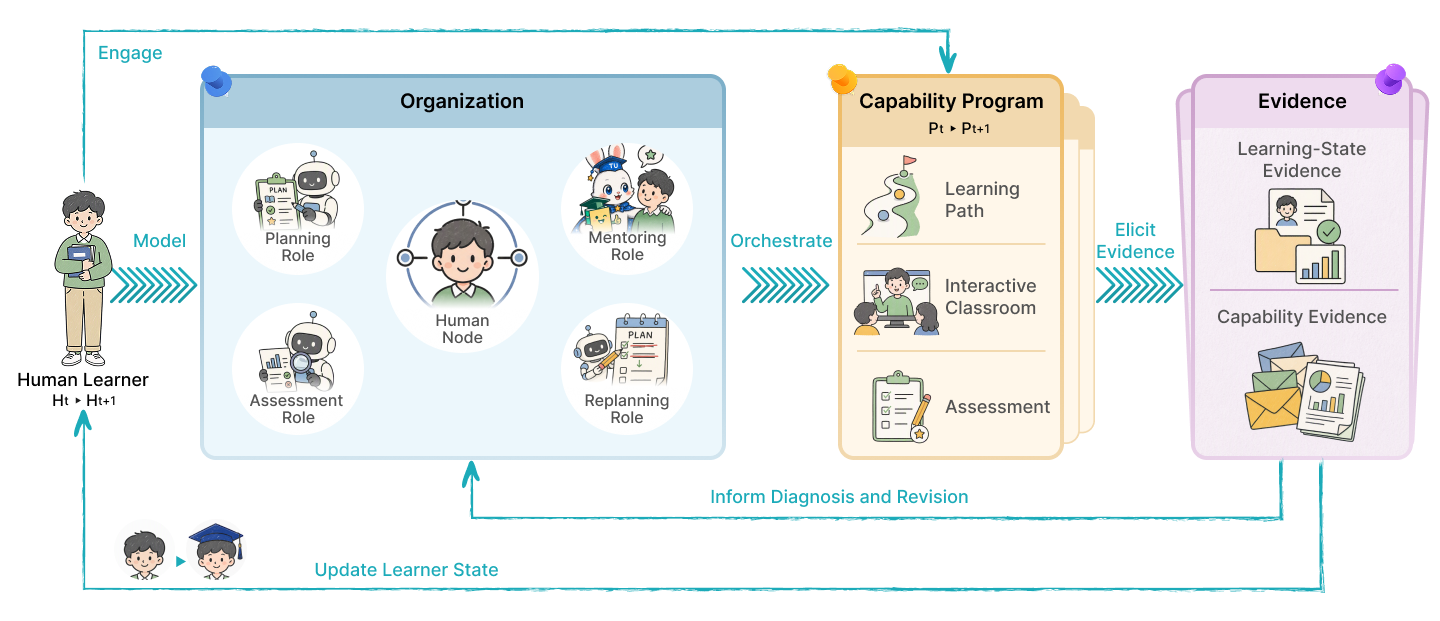}
  \caption{AU connects the Human Node and agent roles through a Capability
  Program. The program includes a Learning Path, Interactive Classroom,
  Assessment. Classroom contributions and assessed
  performance provide Learning-State Evidence and Capability Evidence.
  The agents use this evidence to update learner state and revise remaining work. The four icons denote functional roles; the replanning role includes three diagnostic agents.}
  \Description{The diagram connects the Human Node and agent roles to a
  Capability Program. Classroom and assessment activities produce two types
  of evidence. Both types feed into learner-state updates and replanning.}
  \label{fig:tu-loop}
\end{figure*}

Sustaining this experience across stages requires planning, mentoring, and
assessment to stay aligned with what the learner is trying to achieve. AU
maintains this alignment through two shared representations: the Human Node,
which carries the learner's goals and contributions, and the Capability
Program, which organizes learning activities and assessment criteria around
them (Figure~\ref{fig:tu-loop}). The rest of this section develops the
organization behind the walkthrough: the organizational model
(Section~\ref{sec:tu-organization}), how a Capability Program is formed
(Section~\ref{sec:tu-program}), how executing it produces evidence
(Section~\ref{sec:tu-evidence}), and how that evidence updates learner
state and revises the program (Section~\ref{sec:tu-adaptation}).

\subsection{Organizational Model and Human Node}
\label{sec:tu-organization}
\label{sec:tu-human-node}

AU coordinates work across activities through clear agent
responsibilities and shared learner information.
At time $t$, we represent the organization as
\begin{equation}
\mathcal G_t=(\mathcal V_t,\mathcal R_t),
\qquad
\mathcal V_t=\{H_t\}\cup\mathcal A.
\end{equation}
Here, $H_t$ is the Human Node, and $\mathcal A$ contains role-level Agent Nodes, each representing one or more agents with a shared functional responsibility.
The relations $\mathcal R_t$ specify how nodes share information
and propose or make decisions.

\paragraph{Human Node.}
We define the Human Node as
\begin{equation}
H_t = \langle p_t,\, s_t,\, m_{\leq t} \rangle.
\end{equation}
The profile $p_t$ records the learner's background, experience,
preferences, available time, and practical constraints.
The current state $s_t$ records goals, progress, capability estimates,
difficulties, and uncertainty.
The shared memory $m_{\leq t}$ retains selected evidence, earlier
goals, learner corrections, previous support, and program revisions.

Learners shape this account directly: they state goals, ask questions,
complete tasks, provide feedback, and correct system
assumptions~\cite{bull2007student}. Agents then act on the state these
contributions produce. Assessment results, in turn, ground capability
estimates in the tasks and conditions actually tested.

\begin{figure*}[t]
  \centering
  \includegraphics[width=\textwidth]{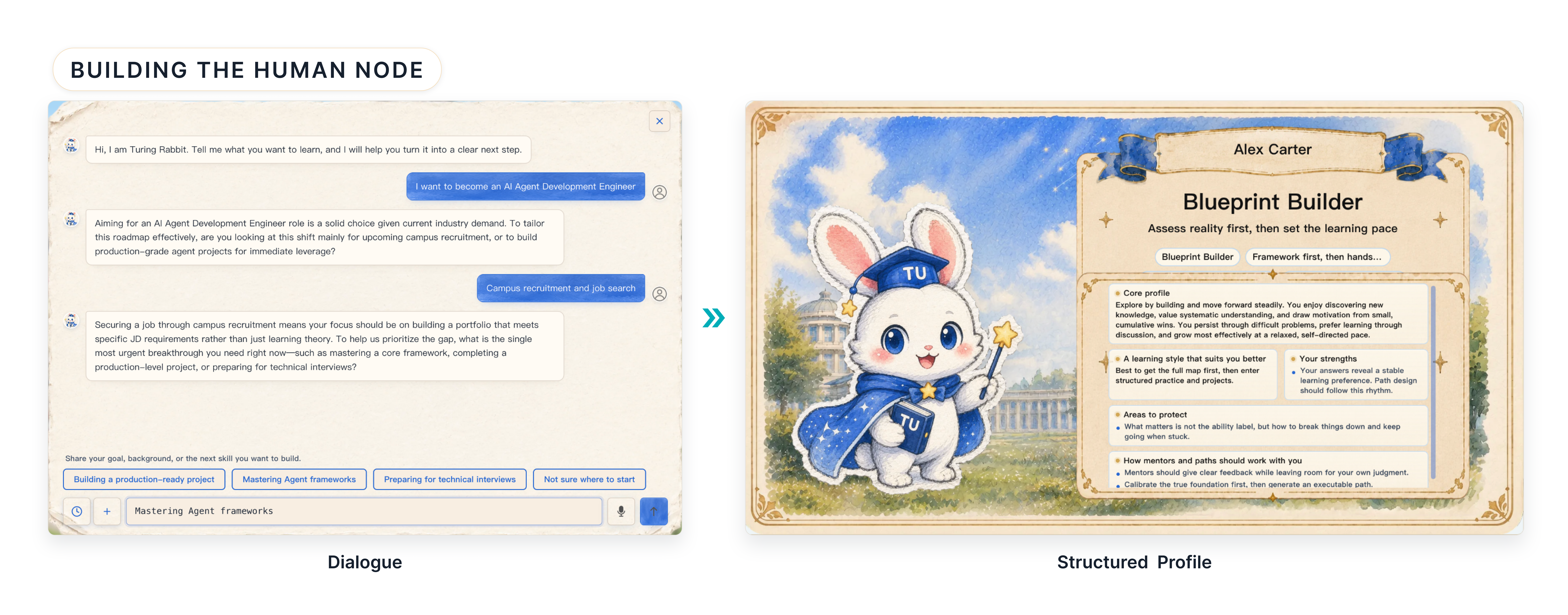}
  \caption{AU uses intake dialogue to build the Human Node. The learner
  describes goals, background, preferences, and constraints. AU stores this
  information in structured records for later planning and interaction. The displayed name and avatar are fictional.}
  \Description{The diagram shows a learner answering intake questions.
  AU turns these answers into structured learner records.}
  \label{fig:tu-human-node-construction}
\end{figure*}

The intake conversation from the walkthrough is where this account begins.
Because a learner's initial request often leaves goals or constraints
underspecified, AU opens with a dialogue that turns the request into
structured goal and profile records
(Figure~\ref{fig:tu-human-node-construction}). The learner may describe
a target role, an intended product, or simply an area of interest;
AU asks follow-up questions only when the answers would change the
plan~\cite{horvitz1999principles,allen1999mixed}. Explicit learner statements
take priority over inferred or default values, whose sources are kept on
record. The resulting records distinguish prior exposure from demonstrated
capability, and project preferences from target criteria, with each
capability claim linked to its evidence and criteria. The planning role
reads these records to set the program's starting point.

\paragraph{Agent Nodes.}
AU represents four functional roles as Agent Nodes,
$\mathcal{A} = \{A_{\mathrm{plan}}, A_{\mathrm{mentor}}, A_{\mathrm{assessment}}, A_{\mathrm{replan}}\}$:
\begin{itemize}
  \item The planning role $A_{\mathrm{plan}}$ uses the learner's goal, state, and
  constraints to form a learning path.
  \item The mentoring role $A_{\mathrm{mentor}}$ creates learning materials and
  provides explanations, practice support, and feedback~\cite{kraemer2010personalizing,kim2006socialcognitive,lester1997persona}.
  \item The assessment role $A_{\mathrm{assessment}}$ creates tasks and criteria
  at the node, stage, and program levels.
  \item The replanning role $A_{\mathrm{replan}}$ uses learner state, program state,
  and evidence to propose changes to remaining work.
\end{itemize}
These roles describe organizational responsibilities, and each role can involve multiple specialized agents. The mentoring role can span domain, career, ethics, platform, and learning-support mentors. The replanning role is implemented by the Replanner, a multi-agent subsystem that coordinates the diagnostic evaluators described in Section~\ref{sec:tu-adaptation} and uses their aggregated diagnosis to guide program revision.

What binds the nodes together is a shared Capability Program $P_t$, which
connects goals, activities, assessment requirements, progress, and revisions.
Each agent consults the Human Node and the current program to decide what
work should follow. We describe this program next.

\subsection{Forming the Capability Program}
\label{sec:tu-program}
\label{sec:tu-path}

\begin{figure*}[t]
  \centering
  \includegraphics[width=\textwidth]{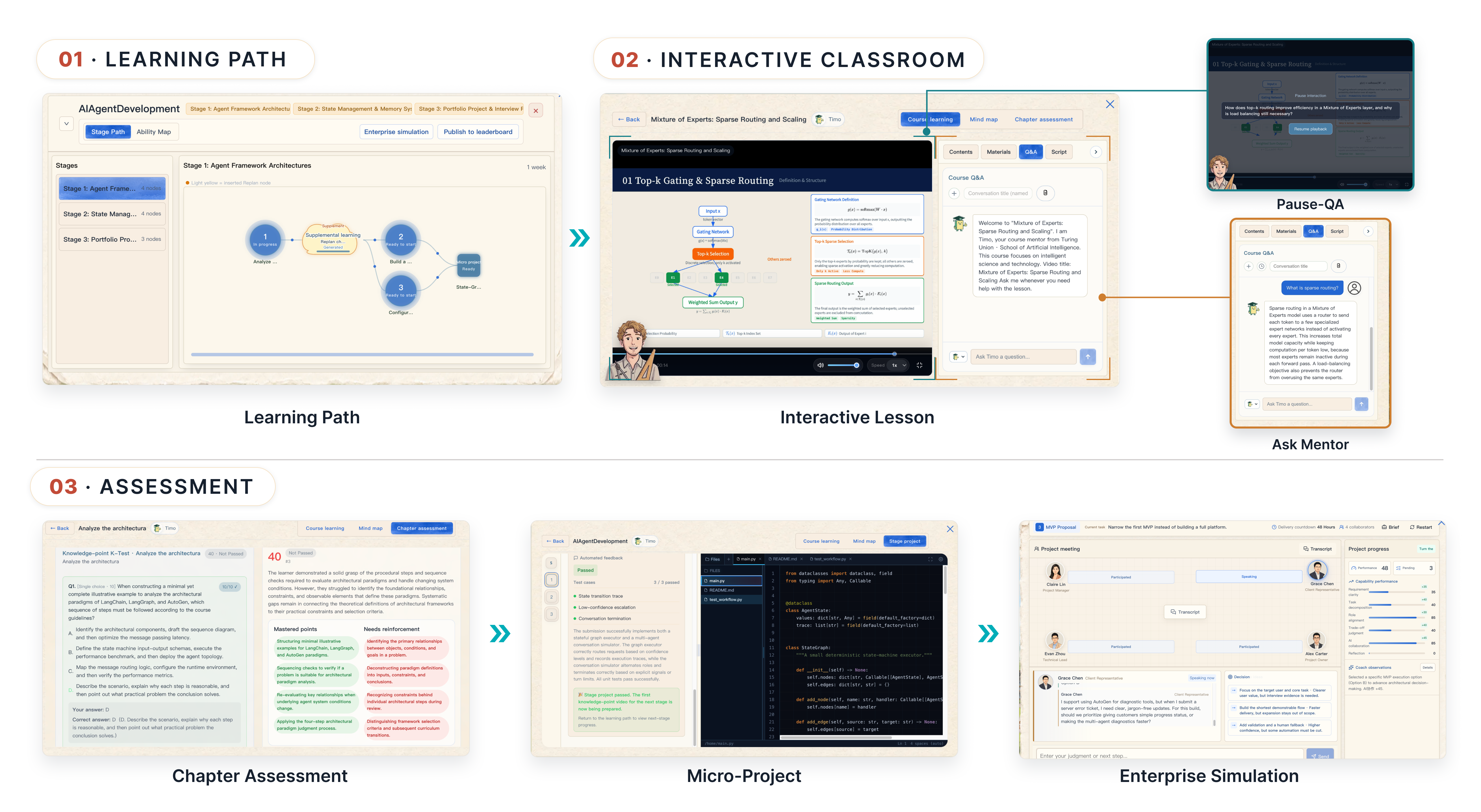}
  \caption{AU breaks the learning goal into stages and linked knowledge-point
  nodes. The classroom supports each node through learning materials,
  Pause-QA, and mentor dialogue. Chapter assessments, stage micro-projects,
  and an enterprise simulation examine performance at increasing scopes. Names and avatars shown in the enterprise simulation are fictional and do not represent actual participants.}
  \Description{The diagram connects stages of knowledge-point nodes to an
  Interactive Classroom. Three assessment levels cover individual knowledge
  points, stage tasks, and a final workplace simulation.}
  \label{fig:tu-capability-program-flow}
\end{figure*}

A learning goal becomes actionable only once its activities and assessment
criteria are pinned down. The \emph{Capability Program} is AU's device for
holding these elements together:
\begin{equation}
P_t=\langle L_t,C_t,Q_t,S_t^P\rangle.
\end{equation}
The Learning Path $L_t$ sets activities and their order; the Interactive
Classroom $C_t$ provides materials and mentor interaction; the Assessment
structure $Q_t$ defines tasks, conditions, and criteria; and the Program
State $S_t^P$ records progress and revisions.
Figure~\ref{fig:tu-capability-program-flow} shows how these components
connect. This subsection covers how the path and program state are formed;
Section~\ref{sec:tu-evidence} turns to the classroom and assessments through
which the program is experienced.

\paragraph{Learning Path construction.}
The planner uses the learner's goal to define stage subgoals and
knowledge-point nodes. It links these nodes through prerequisite
relations within and across stages:
\begin{equation}
L_t=\langle V_t^L,E_t^L,G_t\rangle.
\end{equation}
The set $V_t^L$ contains the nodes grouped into stages. The set $E_t^L$
contains prerequisite links. The conditions $G_t$ govern task
availability and progression.

The planner combines the Human Node with career reference paths,
capability graphs, and job descriptions to identify target skills
and their prerequisites. The learner's background and current state
guide the starting point and prerequisite learning. Preferences and
constraints guide workload, examples, and activity format.

Each node connects its objective to learning materials, mentor support,
and assessment requirements. These connections guide the creation of
lecture documents, videos, Pause-QA activities, and node assessments.
Stage subgoals guide micro-projects, and the overall goal guides the
enterprise simulation. Node objectives and prerequisite links also
provide a reference for later revision. They help the Replanner locate
the affected knowledge point and determine where additional learning
fits within the remaining path.

Execution of the path is tracked through
\begin{equation}
S_t^P=\langle ActivePathVersion_t,Progress_t,Evidence_t,RevisionHistory_t\rangle,
\end{equation}
which identifies the current path, completed and remaining work, assessment
results, and past changes, so that every role works from the same program
state across activities.

\subsection{Executing the Program: Instruction, Assessment, and Evidence}
\label{sec:tu-evidence}
\label{sec:tu-classroom}
\label{sec:tu-assessment}

Once the program is underway, nearly everything the learner does leaves a
trace: a question in a lesson, a retry after feedback, a scored task. AU
treats these traces as evidence, but not all of them warrant the same
conclusions. This subsection follows the program's execution, from
instruction through assessment, to show where evidence comes from and what
each kind can support.

\paragraph{Interactive Classroom and mentor coordination.}
Learning difficulties often arise while learners are studying a knowledge
point, before they reach an assessment. AU therefore attaches an
Interactive Classroom to each node to support learners during this process.
The classroom brings together AU-generated instructional videos~\cite{wang2026teachmaster}, lecture
or session documents, questions, and mentor dialogue, all organized around
the node's objective. Within a lesson, \emph{Pause-QA} inserts questions at
key teaching points, prompting learners to reflect on the ideas just
covered and check their understanding~\cite{roediger2006test}. Learners can
also seek clarification, examples, hints, or feedback on an attempted
solution from mentors~\cite{aleven2016help}. AU selects a mentor by matching
the request against role scope and conversation context. The selected
mentor then responds in light of the node objective and relevant learner
state~\cite{graesser2005autotutor}. These interactions connect explanation,
reflection, and feedback within each lesson, while the resulting questions,
retries, and confidence reports help AU interpret learning difficulties
and support needs.

\paragraph{Layered assessment.}
Understanding a knowledge point is one demand; using knowledge across tasks
is another. AU separates these demands into three assessment levels that
examine understanding, application, and integration in turn:
\begin{itemize}
  \item A \emph{chapter assessment} follows each knowledge-point node.
  It checks the learner's understanding of that point and records grading
  details for each answer.
  \item A \emph{micro-project} follows each stage. It asks the learner to
  apply knowledge from several nodes in a task with defined outputs and
  assessment criteria~\cite{blumenfeld1991motivating,barron1998doing}.
  \item An \emph{enterprise simulation} follows the learning stages for
  the overall goal. It asks the learner to combine knowledge across stages
  in a simulated workplace task~\cite{gulikers2004framework,barnett2002when}.
\end{itemize}
Every assessment specifies the task, the assessed dimensions, the allowed
support, and the grading criteria~\cite{mislevy2003brief}. Where code tasks
support testing, project reviews can combine model feedback with execution
results. Together, the three levels trace a path from local understanding
to broader use of knowledge, and their results feed back at matching scopes:
practice on a knowledge point, review within a stage, or further work on
combining skills.

\paragraph{From interaction to evidence.}
A stated preference and a scored task justify different decisions, so AU
links each observation to the updates it can support. At the framework
level, an evidence item is
\begin{equation}
e_i=\langle x_i,s_i,c_i,t_i,q_i,\lambda_i\rangle,
\end{equation}
recording the observation $x_i$, its source and conditions $s_i$, the
relevant claim or decision $c_i$, the time $t_i$, a confidence or quality
rating $q_i$, and the permitted updates $\lambda_i$. This tuple is the
evidence model that runs through all of AU's records.

The two activities above produce the two kinds of evidence introduced
earlier. Classroom interactions become \emph{Learning-State Evidence} once
AU links them to a learner-state field or teaching decision; this evidence
informs difficulties, preferences, strategies, progress, and uncertainty,
and so guides explanations and support. Assessed performance provides
\emph{Capability Evidence}, which updates capability estimates, but only
within the assessed dimensions and assistance
conditions~\cite{messick1995validity}. A hint request, for example, can
guide support, whereas a scored task can back a claim about the tested
knowledge. Learner corrections update the relevant goal or profile fields
directly.

\subsection{Closing the Loop: State Updates and Program Revision}
\label{sec:tu-adaptation}

AU uses learning evidence to update its account of the learner and
propose changes to the remaining program. We describe the update as
\begin{equation}
(H_t,P_t)\xrightarrow{\mathcal E_t}(H_{t+1},P_{t+1}).
\end{equation}
The evidence set $\mathcal E_t$ informs learner-state updates and
revision proposals. Path changes take effect after learner confirmation.

\paragraph{State update rules.}
The \emph{add} rule combines compatible evidence to strengthen or
weaken an estimate. The \emph{replace} rule updates a goal, preference,
or constraint when its source permits the change and retains its
history. The \emph{keep conflicts} rule preserves conflicting
observations for later diagnosis. For example, high confidence and
repeated errors can remain recorded together. These rules operate
over evidence records, source-linked goal updates, versioned summaries,
and Replan records. Supported gaps guide additional support, while
uncertain or missing evidence calls for further diagnosis or assessment.

\paragraph{Multi-agent diagnosis.}
The Replanner coordinates specialized evaluators to identify which
learning requirements need further support. The evaluators share the
learner's evidence and a node-specific target profile and rubric covering
concept, transfer, and independence dimensions. The evidence includes
assessment results, answer-level errors, task outcomes, reported
difficulties, and help-seeking interactions. Capability Evidence grounds
estimates in assessed performance and its assistance conditions, while
Learning-State Evidence helps interpret difficulties and support needs.

The diagnostic panel comprises three separately prompted LLM evaluators.
An evidence auditor checks the evidential support for judgments, a process
observer examines learning behavior and help-seeking, and an outcome
examiner assesses task performance and errors. Each evaluator assesses
all three dimensions, providing complementary perspectives on the same
evidence under a shared rubric. Disagreement beyond a preset tolerance
in any dimension triggers bounded peer review, during which the
evaluators examine one another's reasoning and reconsider their estimates.
The Replanner aggregates their final estimates using weights obtained
by normalizing each evaluator's self-reported confidence across the panel.
The resulting diagnosis links gaps between estimated performance and
target requirements to the affected node and the learning requirements
that need support.

\paragraph{Diagnosis-guided path revision.}
The revision policy interprets the diagnosed gaps together with
prerequisite performance and earlier interventions to determine whether
a path change is warranted. When the largest decision-adjusted gap falls
below the intervention criterion, AU retains the current path and records
that no revision is needed. When intervention is warranted, the diagnosis
determines both the content and placement of additional learning.
A concept gap guides prerequisite instruction before the affected node,
while a transfer or independence gap guides practice after it.
The diagnosis also specifies the added activity's objective,
instructional content, and assessment requirements, connecting the
identified difficulty to concrete learning tasks and completion criteria.

\begin{figure*}[t]
  \centering
  \includegraphics[width=\textwidth]{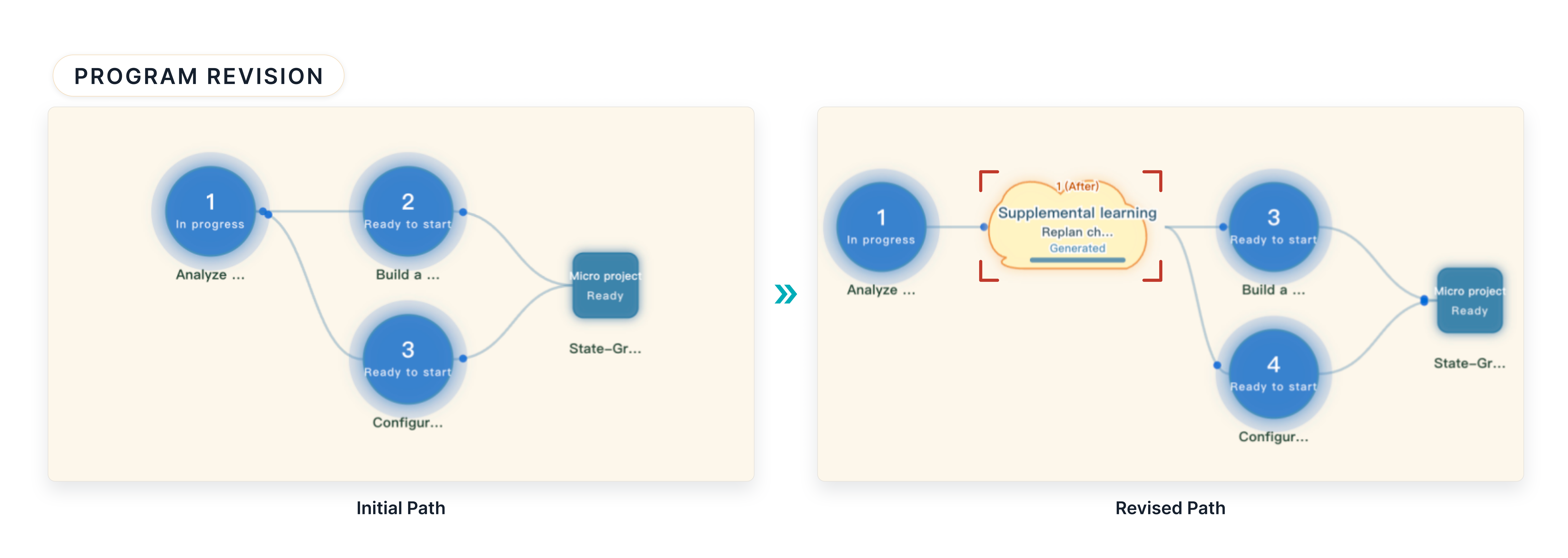}
  \caption{AU uses learner evidence to propose a local path revision.
  The Replanner identifies a supported difficulty and defines additional
  learning within the existing sequence. AU applies the revision after
  learner confirmation and records the updated program state.}
  \Description{The diagram compares a path before and after replanning.
  The revised path adds an activity around the affected node and keeps
  the surrounding learning sequence connected.}
  \label{fig:tu-replan-before-after}
\end{figure*}

The Replanner integrates the proposed activity into the existing
prerequisite sequence, subject to limits on branch depth and insertions
per affected node (Figure~\ref{fig:tu-replan-before-after}).
The revision preserves completed work and the sequence outside the
revised portion, with the original subsequent nodes retained after the
added activity. This local revision policy connects the diagnosed gap
to additional support while preserving the surrounding program structure.
Stage-level cases and broader goal changes follow separate review
procedures.

\paragraph{Learner confirmation and follow-up.}
AU asks the learner to confirm a proposed path change and applies it
after receiving approval. Each Replan record links the supporting
evidence to the diagnosis, selected adjustment, changed and retained
activities, application status, and path versions. Later assessment
results inform follow-up judgments of the learner's understanding and
the outcome of the added support. The revision history provides context
for subsequent teaching and replanning decisions.

\section{Evaluation}
\label{sec:evaluation}

\subsection{Evaluation Overview}
\label{sec:evaluation-overview}

We evaluated AU's support for program formation, evidence-based
revision, and learning during system use.
The three studies assessed initial program quality, revision quality,
and learning outcomes and experiences.
Table~\ref{tab:evaluation-design} links each study to its research
question and evaluation focus.

\begin{table*}[t]
  \centering
  \caption{Research questions and evaluation focus of the three studies.}
  \label{tab:evaluation-design}
  \small
  \setlength{\tabcolsep}{5pt}
  \renewcommand{\arraystretch}{1.15}
  \begin{tabular*}{\textwidth}{
    @{\extracolsep{\fill}}
    p{0.10\textwidth}
    p{0.55\textwidth}
    p{0.27\textwidth}
    @{}
  }
    \toprule
    \textbf{Study} &
    \textbf{Research question} &
    \textbf{Evaluation focus} \\
    \midrule

    Study 1 &
    \textbf{RQ1: Representation.}
    What aspects of learning-program quality does the Capability
    Program representation support? &
    Quality of initial learning programs. \\
    \addlinespace

    Study 2 &
    \textbf{RQ2: Evidence conditioning.}
    How effectively can learner evidence guide program revisions
    that improve learning support and maintain a coherent path? &
    Quality of program revisions based on learner evidence. \\
    \addlinespace

    Study 3 &
    \textbf{RQ3: Human participation.}
    What learning outcomes and experiences arise when learners
    participate in a Capability Program? &
    Learning outcomes, system experience, and experienced revision. \\

    \bottomrule
  \end{tabular*}
\end{table*}

\subsection{Study 1: Capability Program Formation}
\label{sec:study1}

Learning programs need clear activities and criteria for assessing progress.
AU links learning goals to activities, assessments, and success
criteria. Study~1 examines the quality of programs generated with
this approach. The study addresses RQ1 by comparing AU with three
baselines. We focus on how AU organizes learning activities and
defines evidence for assessing the target capabilities.

\subsubsection{Study Design and Measures}

\paragraph{Learner profiles.}
We constructed 96 simulated learner--goal cases by crossing
three learning goals with 16 profile settings
and two natural-language versions of each combination
($3\times16\times2=96$).
We used controlled augmentation of de-identified learner-profile records
from existing system users to vary goal clarity, learning time frame,
AI expertise, and expertise in the target field.
Each method received the same learner and goal facts through its
own input interface.
We used GPT-5.6 Sol~\cite{openai2026gpt56} to simulate learner
dialogues during profile construction.

\paragraph{Baselines.}
General uses a general-purpose LLM to generate learning plans.
PLPP~\cite{lim2026plpp} creates personalized learning paths;
we adapted its workflow to this setting while preserving its core
personalization logic.
GenMentor~\cite{wang2025genmentor} is a multi-agent learning
system and our closest system-level comparison.

\paragraph{Generation and evaluation.}
All four methods used \texttt{gemini-3-flash-preview}~\cite{google2025gemini3}
to generate one program per case (384 outputs), retaining their
native output structures. We compared complete system configurations.
The same \texttt{fable-5} model~\cite{anthropic2026fable5} judged all
outputs with a shared rubric~\cite{zheng2023judging,liu2023geval}.
Two human experts independently rated AU and GenMentor outputs for
24 matched cases, with eight cases per learning goal, after jointly
reviewing the rubric. We averaged their ratings and then averaged
the four primary outcomes to obtain composite scores for each case
and method. Across the resulting 48 case--method pairs, human and
model composites showed Spearman's $\rho=.721$ and Lin's concordance
correlation coefficient of $.708$. These statistics describe pooled
correspondence across the two systems.
Study~2 used the same models in the same roles.

\paragraph{Measures.}
Capability Coverage assesses coverage of the core capabilities
required by the target role.
Program Quality assesses subgoal decomposition, learning unit
granularity, prerequisite coherence, and supporting artifacts.
Assessment Validity assesses whether formative checks, applied
practice, and integrated tasks are designed to elicit evidence of target
capabilities~\cite{mislevy2003brief}.
It also assesses the clarity of success criteria.
Learner Personalization assesses how well the program's
starting point, content, and difficulty fit the learner.
For Capability Coverage, the judge compared expert-annotated Core
Node labels with each program's ordered, flattened learning-unit
titles. Only these title fields were provided for this measure.
Each required node was marked as fully covered, partly covered,
or missing. General and PLPP used a separate, non-scoring
title-extraction step. Program Quality, Assessment Validity, and
Learner Personalization used four, four, and two submetrics,
respectively. Each submetric was rated from 1 to 5. We averaged
submetric ratings within each outcome and rescaled them to the
0--100 reporting scale.

\paragraph{Statistical analysis.}
We report scores on a scale from 0 to 100.
Each case was a unit of analysis.
Wilcoxon signed-rank tests compared AU with each baseline on the
four primary outcomes. We applied Holm correction to these comparisons.
We used bias-corrected and accelerated (BCa) 95\% confidence intervals
for the mean paired differences.

\subsubsection{Higher Ratings for Program Organization and Assessment}
\label{sec:study1-consistency}

\begin{figure*}[t]
  \centering
  \includegraphics[width=\textwidth]{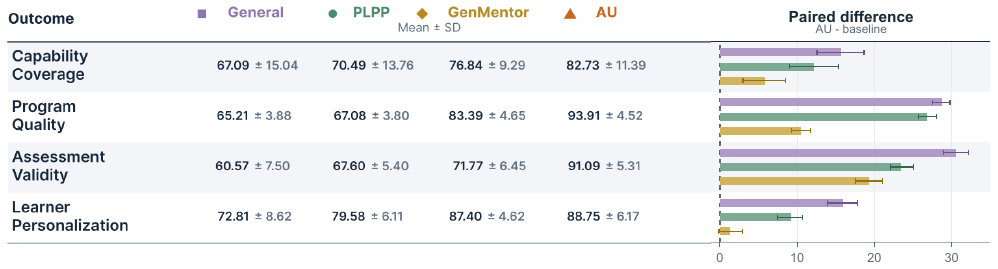}
  \caption{
    Comparison of AU and three baselines across four outcome dimensions. The table reports mean $\pm$ standard deviation (SD). Horizontal bars show mean paired differences in score points for AU minus
    each baseline. Whiskers show BCa 95\% confidence intervals. Purple, green, and gold bars denote comparisons with General, PLPP, and GenMentor, respectively. The four bar groups align with the table rows. The dashed vertical line marks zero difference; positive values favor AU.
  }
  \label{fig:study1-paired}
  \Description{
    The figure shows four primary outcomes for all four methods
    and a grouped bar chart of twelve paired comparisons. AU's largest differences from
    GenMentor concern Assessment Validity and Program Quality.
    The AU--GenMentor interval for Learner Personalization includes zero.
  }
\end{figure*}

AU had the highest mean on all four measures.
GenMentor had the highest baseline mean on each measure.
The comparisons below focus on AU's results against GenMentor.
We also used GenMentor in Studies~2 and~3 because of these results
and its support for revision and interactive learning.

AU's largest advantage was in Assessment Validity.
Its mean was 19.32 points higher than GenMentor's
(95\% CI $[17.55,21.04]$).
AU's Program Quality mean was 10.52 points higher
($[9.27,11.77]$).
Its Capability Coverage mean was 5.89 points higher
($[2.99,8.50]$).
All three comparisons remained significant after Holm correction
(Figure~\ref{fig:study1-paired}).
The largest differences concern the organization of learning tasks
and the design of assessments.
These features specify what learners should do and how their work
will be assessed. AU scored 88.75 on Learner Personalization.
GenMentor scored 87.40.
The paired difference was 1.35 points
(95\% CI $[-0.21,2.92]$, $p_{\mathrm{Holm}}=.090$).

Against GenMentor, AU received higher Program Quality ratings in
85 of the 96 cases and higher Assessment Validity ratings in
94 cases. These counts show that its higher ratings extended
across most matched cases.

\subsubsection{Applied Tasks and Clear Success Criteria}

\begin{figure*}[t]
  \centering
  \includegraphics[width=\textwidth]{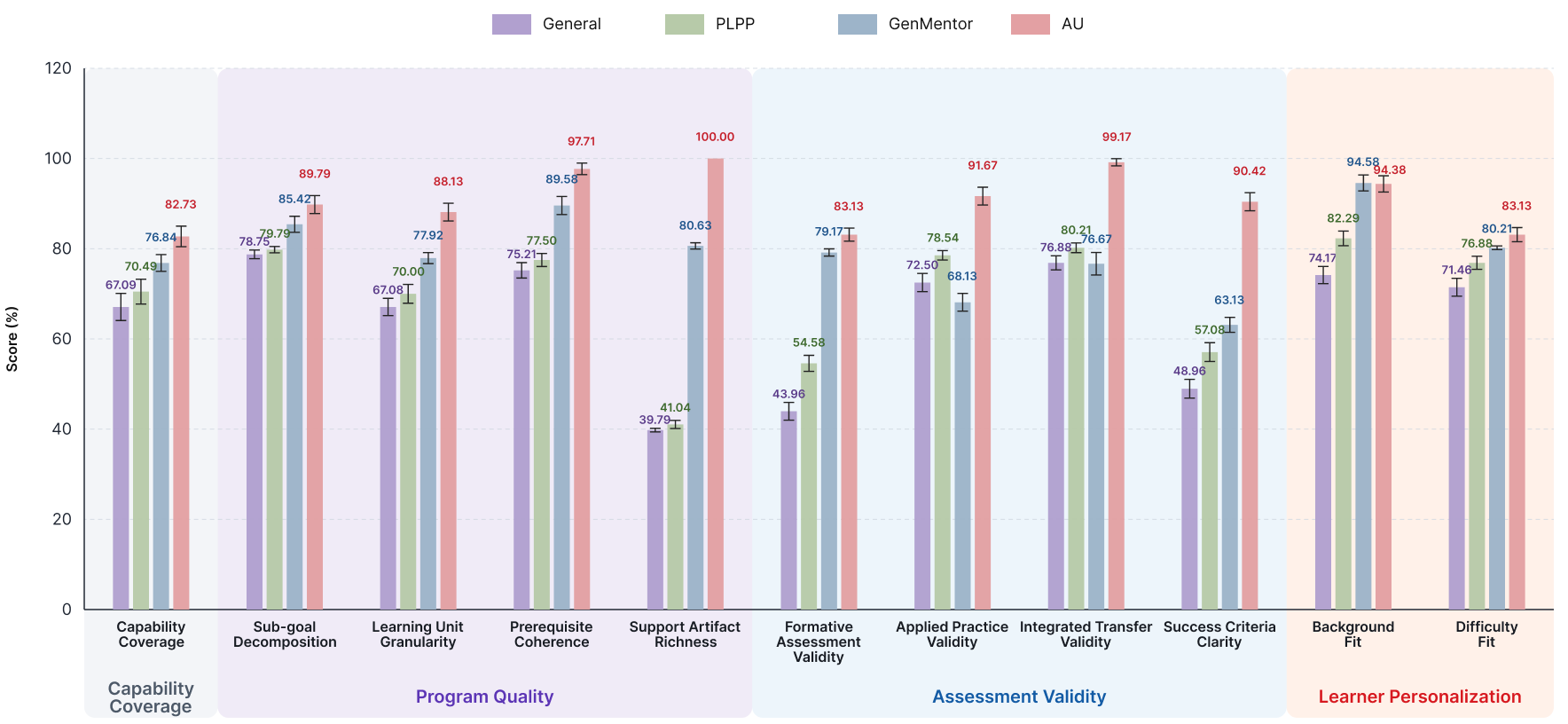}
  \caption{
    The figure shows Capability Coverage and ten rubric submetrics
    for the four methods. Bars show means.
    Whiskers show 95\% confidence intervals across 96 matched cases.
    Shading groups the submetrics by their main measure.
    The figure shows Capability Coverage separately.
  }
  \label{fig:study1-submetrics}
  \Description{
    The figure compares all four methods on Capability Coverage
    and ten submetrics. AU has large mean advantages in Applied
    Practice Validity, Integrated Transfer Validity, and Success
    Criteria Clarity.
  }
\end{figure*}

The submetric scores locate AU's higher ratings within task design
(Figure~\ref{fig:study1-submetrics}).
AU exceeded GenMentor by 23.54 points on Applied Practice Validity
and 22.50 points on Integrated Transfer Validity.
These ratings concern tasks that ask learners to apply and
combine the target capabilities.
AU also scored 27.29 points higher on Success Criteria Clarity.
Its programs provided clearer standards for judging task outcomes.
These standards define what a completed task should achieve
and give learners a concrete target for their work.

The systems organized their outputs at different levels.
General and PLPP mainly produced plan text.
GenMentor added documents and quizzes for each session.
AU linked teaching materials, assessments, applied tasks, and
integrative projects across capability, stage, and program levels.
This structure places practice and assessment within the learning
plan. The higher ratings for tasks and success criteria support
the value of this design in the generated programs.

\subsection{Study 2: Evidence-Conditioned Program Evolution}
\label{sec:study2}

Study~1 examined programs built from learners' starting profiles.
New assessment results and questions can reveal where learners
need more support. AU uses this evidence to revise the remaining
program. Study~2 addresses RQ2 by assessing how well these revisions
improve program fit and maintain path coherence.

\subsubsection{Study Design and Measures}

\paragraph{Cases and conditions.}
We analyzed 32 source cases from Study~1 associated with the target role of “AI Agent Development Engineer.” Each case was evaluated at two points along the learning trajectory (early vs. late). For each trajectory position, we specified two target assessment scores (20/100 and 50/100), yielding a fully crossed design with $32\times2\times2=128$ matched case pairs for AU and GenMentor. The two trajectory positions operationalized differing amounts of remaining learning, whereas the two target scores represented distinct levels of low assessment performance. This design allowed us to examine whether revision behavior remained appropriate under both severe and moderate learning difficulties. Each system maintained its own initial programs and assessment items. Case matching controlled for learner profile, trajectory position, and target score. All experiments employed the model configuration described in Study~1.

\paragraph{Evidence and revision.}
For AU, the learner simulator also asked questions about difficult
content. AU answered through its normal tutoring process.
AU retained these dialogues as conversational evidence alongside
the assessment results. AU's normal trigger initiated the Replanner's multi-agent diagnosis, which informed its local revision policy (Section~\ref{sec:tu-adaptation}). A script sent Replan requests to GenMentor.
Each system revised its program using the evidence available
through its own workflow. The shared external judge evaluated each revision using the original program, revised program, and case context.

\paragraph{Measures.}
We used four measures to assess program improvement, path coherence,
revision decisions, and their supporting evidence. Program Improvement measures the change in program quality
relative to the system's own initial program. It assesses whether
the revised program better fits the learner's current state. Post-Revision Path Coherence measures whether the revised
path maintains prerequisite relationships, coherent stage transitions,
and appropriate difficulty progression. We use this measure to
check whether the added support fits the remaining program. Revision Decision Appropriateness measures whether the
chosen changes address the learner's needs and preserve content
that remains appropriate. This measure assesses the scope and
suitability of the revision. Evidence-Revision Alignment measures whether the available
learning evidence and learner state support the main changes.
This measure assesses the basis for the revision.

\paragraph{Statistical analysis.}
Program Improvement used ratings from $-2$ to $+2$.
The other measures used ratings from 1 to 5.
We report normalized means on a scale from 0 to 100.
We used case-level paired bootstrap 95\% confidence intervals
for the mean differences between AU and GenMentor.
Each of the 32 source profiles appeared in four conditions.

\subsubsection{Higher Revision Quality}
\label{sec:exp2-quality}
\label{sec:exp2-alignment}

\begin{figure*}[t]
    \centering
    \includegraphics[width=0.86\textwidth]{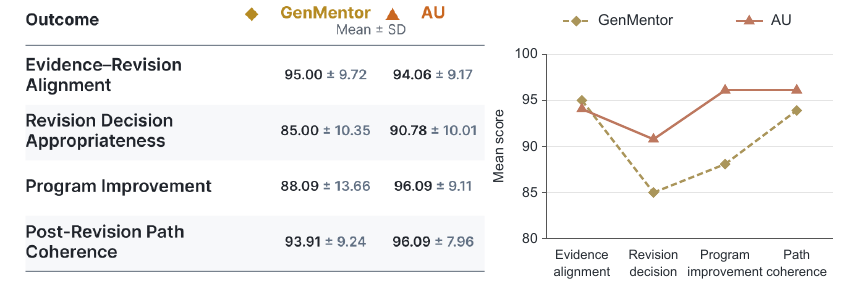}
    \caption{
        The table reports mean $\pm$ standard deviation (SD) on a normalized 0--100 scale. The right panel compares mean scores across the four revision measures. Paired AU $-$ GenMentor mean differences (95\% paired bootstrap confidence intervals), in the plotted order, are $-0.94\,[-3.28,1.56]$, $+5.78\,[3.28,8.28]$, $+8.01\,[5.27,10.74]$, and $+2.19\,[0.31,4.06]$.}
    \label{fig:exp2-revision-quality}
    \Description{
        The table and two line profiles compare AU and GenMentor. AU has higher mean ratings for Program Improvement,
        Revision Decision Appropriateness, and Post-Revision Path
        Coherence. Both systems have high Evidence-Revision
        Alignment ratings.
    }
\end{figure*}

AU received higher mean ratings on Program Improvement,
Revision Decision Appropriateness, and Post-Revision Path Coherence
(Figure~\ref{fig:exp2-revision-quality}).
Its Program Improvement mean was 8.01 points higher than
GenMentor's. This score compares each revision with the system's
own initial program. AU's higher score shows greater rated
improvement in how the program fits the learner's current needs.

AU's Revision Decision Appropriateness mean was 5.78 points higher.
Its Post-Revision Path Coherence mean was 2.19 points higher.
These ratings indicate more appropriate revision decisions and
more coherent resulting paths under the rubric.
All three paired bootstrap intervals were above zero.

Both systems received high Evidence-Revision Alignment ratings.
AU scored 94.06, and GenMentor scored 95.00.
The mean difference was $-0.94$ points
(95\% CI $[-3.28,1.56]$).
These ratings indicate that both systems' revisions were well
supported by the evidence available in their workflows.

\begin{figure}[!t]
    \centering
    \includegraphics[width=0.76\linewidth]{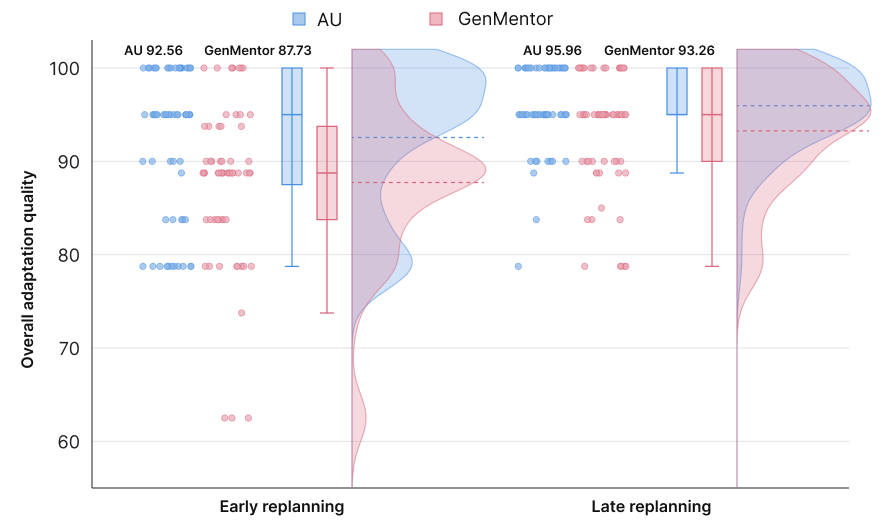}
    \caption{Overall adaptation quality for AU and GenMentor under early
    and late replanning. Points show individual cases.
    Boxplots summarize the distributions, and shaded curves show
    score densities. Dashed lines and labels indicate means.}
    \label{fig:study2-trajectory}
    \Description{
        Overall Adaptation Quality for AU and GenMentor under early
        and late replanning. Points show individual cases, boxplots
        summarize the distributions, and dashed lines indicate means.
    }
    % \vspace{-0.5\baselineskip}
\end{figure}

\paragraph{Revision across the learning trajectory.}
Figure~\ref{fig:study2-trajectory} compares early and late replanning.
The 20/100 and 50/100 conditions were pooled within each position,
yielding 64 cases per system. Overall Adaptation Quality is the
equal-weight mean of the four normalized revision measures, used here
as a summary. It was lower under early than late replanning for both
systems: AU increased from 92.56 to 95.96, while GenMentor increased
from 87.73 to 93.26. The AU--GenMentor difference was larger early
than late (4.83 vs.\ 2.70 points), suggesting a clearer distinction
when more of the program remained. This comparison is descriptive and
does not establish that the two gaps differ statistically.

\subsubsection{Addressing Learning Difficulties with Structured Tasks}
\label{sec:exp2-cases}

Two selected cases illustrate how the systems organized support
in response to their respective assessment evidence.
AU specified remedial activities with explicit learning steps
and completion requirements.
In both cases, AU received $+2$ for Program Improvement and 5/5
for Revision Decision Appropriateness. GenMentor received $+1$
and 4/5 on these measures. Both systems received 5/5 for
Evidence-Revision Alignment and Post-Revision Path Coherence.

\paragraph{Structuring prerequisite learning.}
In CASE-0025 (early, 20/100), GenMentor revised its first
session to address context management, structured output, and
error handling. It added explanations, practice tasks, and guided
problem solving within that session.

AU's assessment identified confusion about prompting,
retrieval-augmented generation (RAG), and agent-based approaches.
AU added a separate activity on these concepts and when to use
each approach.
The activity included positive and negative examples, checks
on when each approach applies, and a small example task.
The plan placed these learning steps and checks before the
return to the original learning node.
AU made prerequisite learning, understanding checks,
and the return to the original node explicit parts of the path.
This design specified how the added support connected to the
remaining program and how learners would complete it.

\paragraph{Defining practice and completion criteria.}
In CASE-0008 (late, 50/100), GenMentor revised its offline
evaluation session. It added explanations and code examples on
retrieval-ranking metrics and model-based evaluation.
These changes addressed weaknesses found in its assessment.

AU addressed difficulties in containerized deployment and
configuration through a separate practice task.
The task required independent completion, error handling,
and explicit recovery procedures and acceptance criteria.
AU's revision specified how learners should use the added
content and check their results.
The task defined both the work learners would perform and the
standards their work should meet.

\subsection{Study 3: Learning Outcomes and Learner Experience}
\label{sec:study3}

A learning program needs to help learners use knowledge, act on
feedback, and continue when they face difficulty. Studies~1 and~2
examined the quality of generated programs and their revisions.
Study~3 addresses RQ3 by examining the value of these supports
during actual learning. We combined learning outcomes, experience
ratings, and interviews to examine changes in performance alongside
learners' reported actions and judgments of their own progress.

All participants followed the assigned goal of becoming an
\emph{AI video generation product manager}. We chose this goal
because it combines AI knowledge with product decisions.
We compared the complete AU and GenMentor workflows.
GenMentor had the highest mean scores among the Study~1 baselines
and supports interactive learning and revision~\cite{wang2025genmentor}.

\subsubsection{Study Design and Procedure}
\label{sec:study3-design} 

\begin{figure*}[t]
  \centering
  \includegraphics[width=\textwidth]{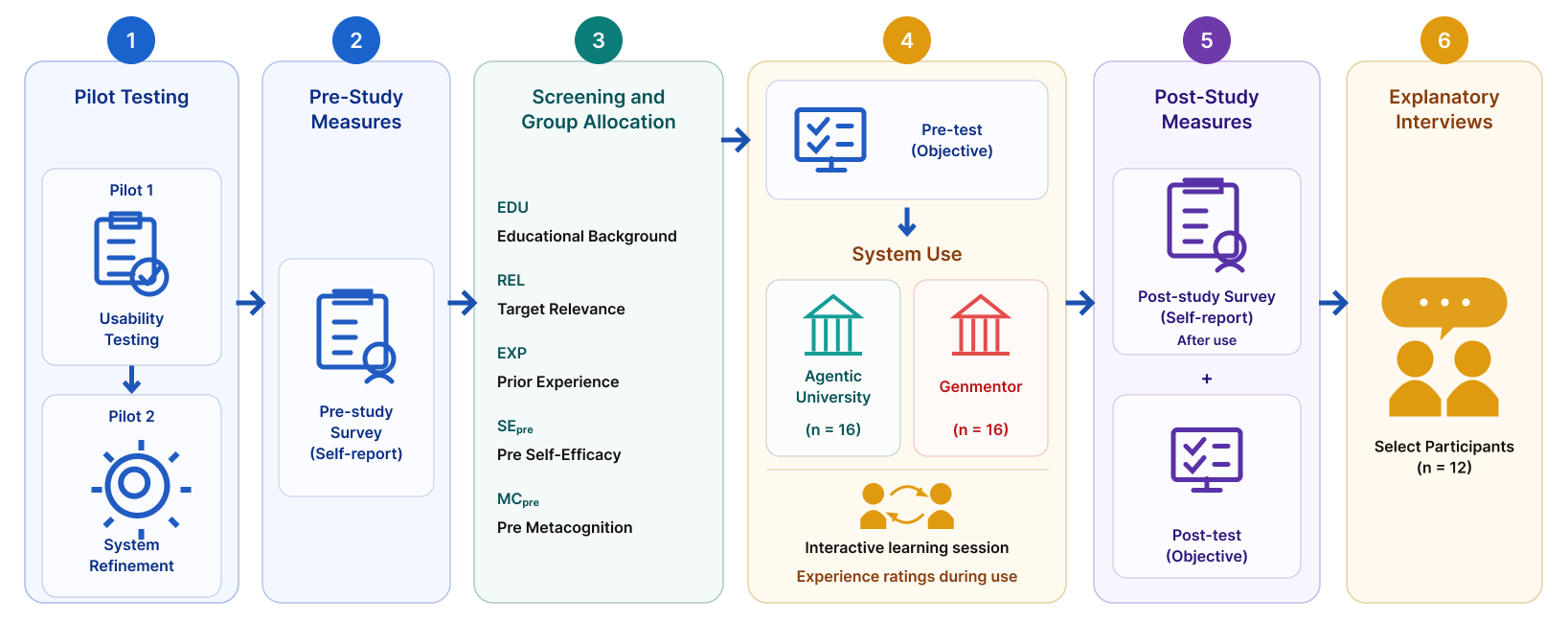}
  \caption{Overview of the Study~3 procedure, including pilot testing,
  participant screening and allocation, pre- and post-study measures,
  system use, and explanatory interviews.}
  \label{fig:study3-procedure}
  \Description{Study 3 procedure from internal pilot testing through
  pre-study measures and screening, allocation of 16 participants to AU
  and 16 to GenMentor, objective pre-testing, system use with experience
  ratings, post-study measures and objective post-testing, and explanatory
  interviews with selected participants.}
\end{figure*}

Figure~\ref{fig:study3-procedure} summarizes the Study~3 procedure,
from pilot testing and pre-study screening through system use,
post-study measures, and explanatory interviews.

\paragraph{Pilot testing.}
The team ran two internal pilot rounds with 30 system developers
and student interns in total. The first round assessed usability.
The second informed system refinement before the main study.

\paragraph{Recruitment and pre-study survey.}
Forty individuals registered through online channels. Six did not proceed because of health issues, scheduling constraints, or ineligibility, leaving 34 who completed the pre-study survey. The survey recorded educational stage, disciplinary relevance, prior domain-related experience, Self-Efficacy, and Metacognition. Before allocation, we excluded two respondents whose baseline Self-Efficacy and Metacognition scores both exceeded 90/100. This study-specific criterion aimed to reduce ceiling constraints on measuring positive changes in self-ratings~\cite{wang2008ceiling}. No score-based eligibility thresholds were applied to the three background indicators. The final sample comprised 32 learners aged 18--30. All participants provided informed consent, and the university's ethics committee approved the study. Participants who completed the study received CNY~100. 

\paragraph{Group allocation and pre-test.}
Before system use, we generated five random allocation schemes and selected the final allocation by considering between-group distributions across the five pre-study indicators, yielding 16 participants in AU and 16 in GenMentor.
Both groups pursued the same learning goal.
We selected ten parent questions from online accounts of
product-manager job interviews and used their decision contexts
to develop the tests. An item-generation agent produced questions,
correct answers, and distractors within the knowledge covered by
both systems. It generated Forms A and B in isolated environments
using the same prompt and configuration. Form A served as the
pre-test and Form B as the post-test; both measured Objective
Capability on a 0--50 scale. Participants completed Form A after
group allocation and before system use.

\paragraph{System use and experience ratings.}
Both groups used the same completion window of approximately
54 hours. The window covered the pre-test, learning activities,
and post-test.
AU organized videos and quizzes into learning units called knowledge
nodes. It also provided short written projects, called microprojects,
and a project simulation through dialogue.
GenMentor organized text materials and quizzes into sessions.
AU used assessment results and learner feedback to identify revision
needs. It requested confirmation and inserted additional learning into the remaining path after learner approval. GenMentor learners clicked the Replan button
to request a new learning path based on session assessment results.

Participants opened the experience questionnaire at the start and
completed relevant items during use. These items assessed
participants' perceptions of learning support.
Instructional Alignment assessed perceived alignment among goals,
content, assessments, and initial knowledge.
Learner Agency \& Engagement assessed perceived support for acting
on feedback, active thinking, and control over learning.
Cognitive Support assessed the clarity and organization of
explanations and materials.
Learning Efficiency measured perceived time
use and ease of task completion.

Participants who reported revision also rated their overall revision
experience. Adaptive Diagnosis covered revision timing and the fit
of the diagnosis to their difficulties.
Instructional Responsiveness covered the fit and continuity of added
instruction. Post-Revision Recovery covered perceived ability to
retry or continue learning.

\paragraph{Post-study measures.}
After completing their learning tasks, participants took the
objective post-test (Form B). They also completed the post-study
Self-Efficacy and Metacognition items and the remaining experience
items. Self-Efficacy used four items on task
confidence~\cite{bandura2006guide}. Metacognition used three items
on self-reported understanding and
monitoring~\cite{flavell1979metacognition,schraw1994assessing}.
All 32 allocated participants completed the study. The matched
pre- and post-study self-ratings used the same item wording and
response scales. We averaged item scores within each multi-item
construct and report self-rating scores on a 0--100 scale.

\paragraph{Explanatory interviews.}
We selected 12 interview volunteers from the study participants,
six per group, to cover different backgrounds, revision experiences,
and patterns of test and self-rating change. Both groups used the
same semi-structured guide.
Interviews were in Chinese and lasted 50 to 75 minutes.
Participants reopened their systems to help recall specific learning
and revision events.
Each interviewee received an additional CNY~50.
We label AU interviewees T1 to T6 and GenMentor interviewees G1 to G6.

\subsubsection{Data Analysis}
\label{sec:study3-analysis}

\paragraph{Quantitative analysis.}
Each participant was one unit of analysis.
Learning gain was the post-study score minus the pre-study score.
We report means with standard deviations (SDs).
We used Wilcoxon signed-rank tests for within-group changes.
We used Mann-Whitney $U$ tests~\cite{mcknight2010mann} to compare
participant-level gains and questionnaire scores between groups.
We report rank-biserial correlations ($r_{rb}$) as effect sizes.
Positive values favor AU.
Holm correction covered the four experience comparisons and the
three revision comparisons as separate families.
Learning-gain comparisons report raw $p$ values.
All tests were two-sided with $\alpha=.05$.

\paragraph{Qualitative analysis.}
The research questions and quantitative results guided the broad
analysis areas. Within these areas, we used \emph{inductive thematic analysis}. We developed codes from participants' explicit accounts
of their experiences~\cite{braun2006using}.
Two researchers coded independently and resolved differences
through discussion. We counted each participant at most once
per theme. The theme counts describe the interview sample.
Italic quotations are English translations of Chinese interview
excerpts. We interpreted the accounts alongside
test scores, questionnaire ratings, and system records of learning
interactions and assessment results.

\subsubsection{Findings}
\label{sec:study3-findings}

\begin{figure*}[t]
  \centering
  \includegraphics[width=\textwidth]{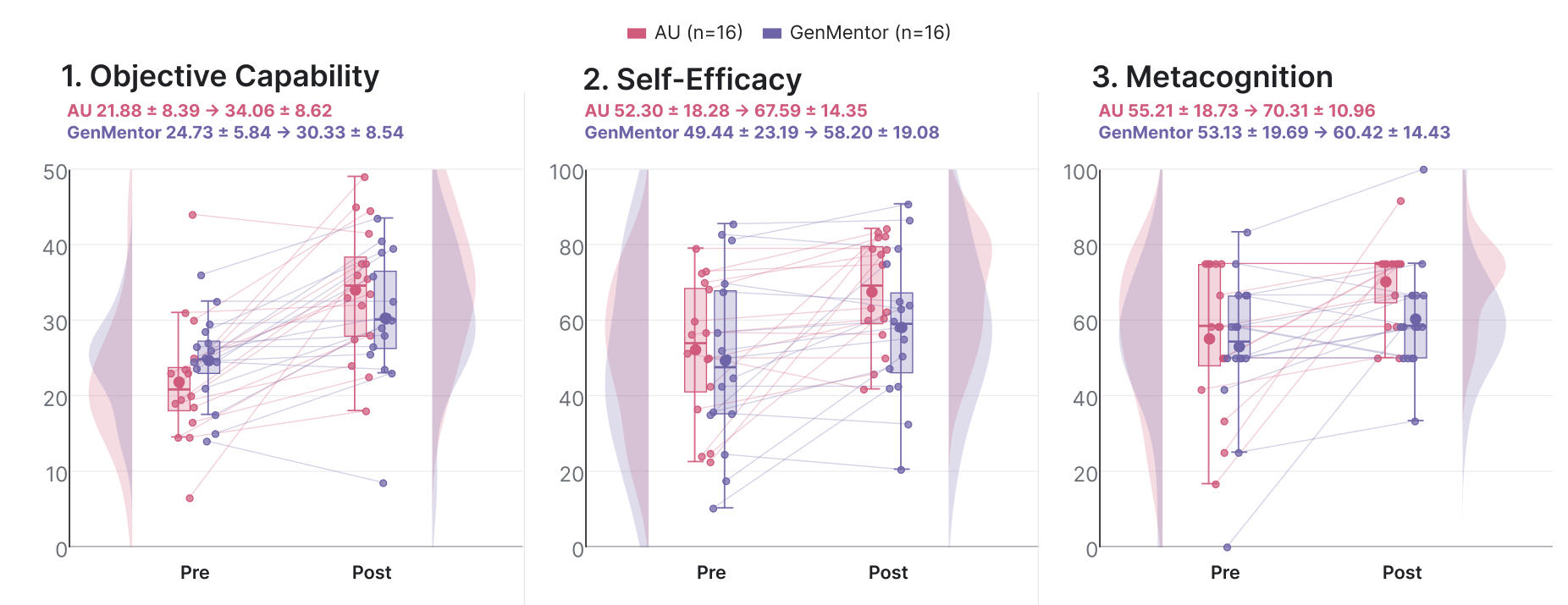}
  \caption{Learning outcomes before and after system use.
    Raincloud plots show Objective Capability, Self-Efficacy, and
    Metacognition for 16 participants in each group. Lines link the
    two observations for each participant. Labels report mean $\pm$ SD.
    Objective Capability uses a scale from 0 to 50. The self-ratings
    use scales from 0 to 100.}
  \label{fig:study3-outcomes}
  \Description{The plots show scores before and after learning for AU
    and GenMentor. AU has larger mean gains on all three outcomes.
    Each line links scores from the same participant.}
\end{figure*}

\paragraph{Larger learning gains and knowledge use in tasks.}
AU showed larger mean gains on all three learning outcomes
(Figure~\ref{fig:study3-outcomes}). Objective Capability increased
from $21.88\pm8.39$ to $34.06\pm8.62$ under AU and from
$24.73\pm5.84$ to $30.33\pm8.54$ under GenMentor. Both groups
improved, with within-group $p<.001$ for AU and $p=.006$ for
GenMentor. Their mean gains were 12.19 and 5.61 points.
AU's mean gain was 6.58 points larger. The gain-score comparison
yielded $U=184.5,p_{\mathrm{raw}}=.035,r_{rb}=.44$.
Self-Efficacy gains were 15.30 and 8.77 points ($p=.300$).
Metacognition gains were 15.10 and 7.29 points ($p=.463$).
AU's clearest observed learning advantage was its larger gain
in assessed knowledge and task judgments. The two self-ratings
showed higher mean gains for AU without a statistically significant
group difference.

Interviewees described how they used new knowledge to answer questions.
T1 recalled guessing on the pre-test. T1 then explained,
\emph{``After learning, I knew the correct answer and why it was
the right choice.''} T5 reported a similar change in questions about
adapting a system to user needs. T5 could understand the task and make
relevant choices after learning. In the GenMentor group, G1
linked limits in video generation to product choices.

Three of the six AU interviewees described reviewing materials or
revising work after feedback (T1, T2, T3). T2 and T3 returned to
materials to check concepts they had missed. T1 described a further
step in the microprojects:
\emph{``I would revise my work based on the feedback and submit
it again.''} T1 linked these open tasks and their feedback to greater
confidence. For T1, the revised output provided a concrete basis
for judging progress. These accounts describe how learners used
feedback to review concepts and revise their work.

Participants also used different standards to judge their learning.
T3 described greater understanding but linked lower confidence during
the study partly to fatigue. G3 judged understanding by the links
between concepts and reassessed earlier knowledge after further review.
These accounts suggest that self-ratings can reflect perceived
learning, fatigue, and the standards learners use to judge progress.

\begin{figure*}[t]
  \centering
  \includegraphics[width=0.86\textwidth]{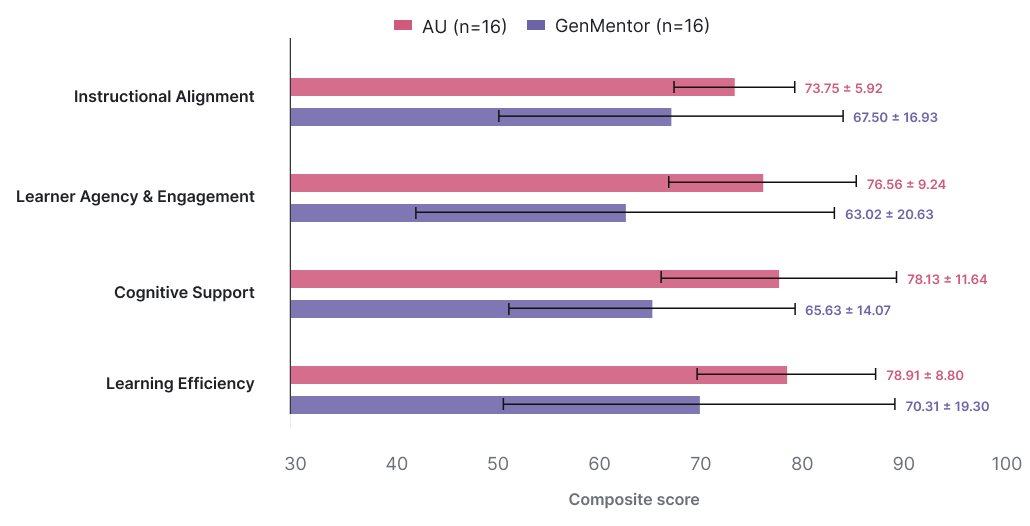}
  \caption{Experience ratings collected during and after system use.
    Bars show the means of four measures. Error bars show $\pm1$ SD.
    Each group includes 16 participants.}
  \label{fig:study3-experience}
  \Description{The plot compares Instructional Alignment, Learner
    Agency and Engagement, Cognitive Support, and Learning Efficiency
    and Interaction Support. AU has higher mean ratings on all four
    measures. The Cognitive Support comparison remains significant
    after Holm correction.}
\end{figure*}

\paragraph{Cognitive support and choices within the learning path.}
AU received higher mean scores on all four experience measures
(Figure~\ref{fig:study3-experience}). Cognitive Support was
$78.13\pm11.64$ under AU and $65.63\pm14.07$ under GenMentor.
The mean difference was 12.50 points. The group comparison remained
significant after Holm correction
($U=190.5,p_{\mathrm{raw}}=.012,p_{\mathrm{Holm}}=.048,r_{rb}=.49$).
Instructional Alignment was 73.75 versus 67.50.
Learner Agency \& Engagement was 76.56 versus 63.02.
Learning Efficiency was 78.91 versus 70.31.
These three comparisons showed higher means for AU. Their Holm-adjusted
$p$ values were .193, .126, and .193.

AU interviewees described support that made the content easier to use.
T1 valued short videos and materials that covered the main concepts
and helped form an overview of the target role. T2 valued video
summaries, AI explanations, and notes:
\emph{``The system could also create notes based on the questions
I asked in the AI conversation.''} These accounts describe concrete
forms of cognitive support: concise materials, accessible
explanations, and resources that learners could revisit.

Four of the six AU interviewees described choices about resources or
learning pace (T2, T3, T4, T6). T4 could replay a difficult section
or look up a concept. T6 moved between video and written materials.
T3 explained,
\emph{``I could decide whether to look at these materials in detail
and whether to talk with the AI model.''} T5 also described questions
within lessons:
\emph{``The lessons often asked us questions and let us think before
continuing.''} Group G had the following feedback: G4 thought \emph{``there's a lack of dwell time, common mistakes, and judgment for short-answer questions; hints mostly appear after submission''}, while G3 felt \emph{``the hint words are pretty experienced, but contextual short answers reveal gaps in knowledge connection and practical application''}. These accounts show that learners could follow AU's
learning path and choose how to work through each part. They could
revisit a difficult section, switch materials, or ask for an explanation.

\begin{figure}[!t]
    \centering
    \includegraphics[width=0.66\linewidth]{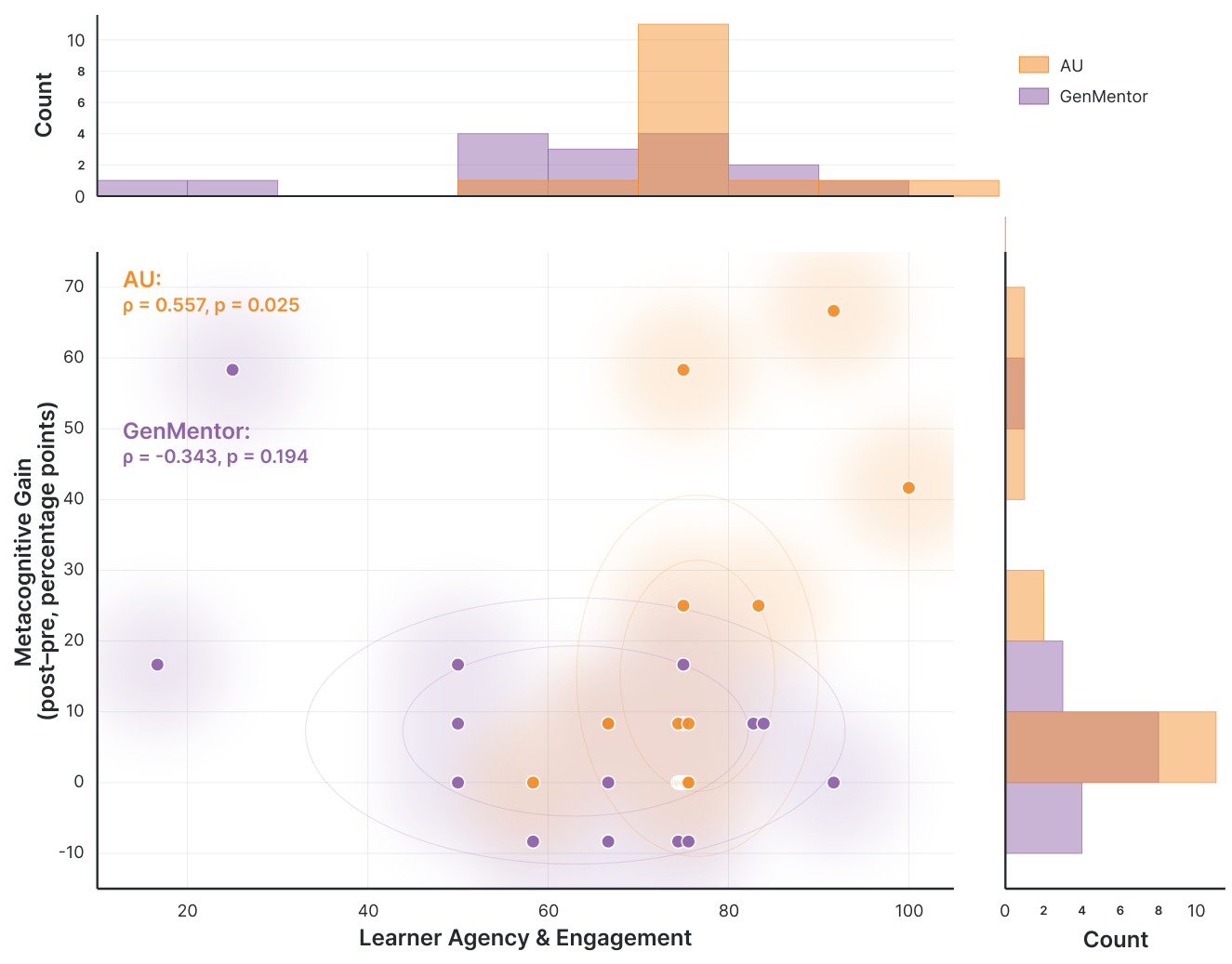}
    \caption{
        Participant-level association between Learner Agency
        \& Engagement and pre--post Metacognition gain.
    }
    \label{fig:study3-agency-growth}
    \Description{
        A scatterplot with marginal histograms compares AU and GenMentor
        participants. The horizontal axis is Learner Agency and Engagement,
        and the vertical axis is pre--post Metacognition gain. AU shows a
        positive association, while GenMentor shows no corresponding
        positive trend.
    }
    % \vspace{-0.5\baselineskip}
\end{figure}

\paragraph{Learner agency and perceived growth.}
Figure~\ref{fig:study3-agency-growth} examines whether learners'
reported agency was associated with their perceived growth during the
study at the participant level. Within AU, stronger Learner Agency
\& Engagement was positively associated with greater pre--post
Metacognition gain ($\rho=.557$, $p=.025$), whereas GenMentor showed
no corresponding positive relationship ($\rho=-.343$, $p=.194$).
Within AU, learners who reported stronger participation and control also tended to report greater growth in metacognitive judgment. This exploratory association was not observed in GenMentor.

\paragraph{Responding to difficulties and supporting the next step.}
Revision-experience ratings came from 13 AU participants and seven
GenMentor participants who reported using the revision function.
AU asked for confirmation when it identified a need for revision.
GenMentor learners started revision through Replan.
The following exploratory comparisons use these revision-experience
subsamples.

Instructional Responsiveness showed the largest descriptive
difference among the revision measures.
AU scored 78.85 and GenMentor scored 64.29, a difference of 14.56 points
($U=72.0,p_{\mathrm{raw}}=.030,p_{\mathrm{Holm}}=.090,r_{rb}=.58$).
Post-Revision Recovery was 75.00 versus 71.43
($p_{\mathrm{Holm}}=.640$).
Adaptive Diagnosis was 74.04 versus 80.36
($p_{\mathrm{Holm}}=.378$).
None of the three exploratory comparisons reached the corrected
significance threshold. Table~\ref{tab:study3-revision} reports the
full results.

%% \FloatBarrier

\begin{table*}[t]
  \centering
  \caption{Revision experience among participants who reported revision.
    Values are mean $\pm$ SD on a scale from 0 to 100. Each participant
    provides one score for each measure across their revision experience.
    Two-sided Mann-Whitney $U$ tests compare the groups. Holm correction
    covers the three comparisons. Positive rank-biserial effect sizes
    ($r_{rb}$) favor AU.}
  \label{tab:study3-revision}
  \small
  \setlength{\tabcolsep}{3pt}
  \renewcommand{\arraystretch}{1.12}
  \begin{tabular*}{\textwidth}{
    @{\extracolsep{\fill}}p{0.24\textwidth}cccccc@{}
  }
    \toprule
    \textbf{Construct}
      & \textbf{AU ($n=13$)}
      & \textbf{GenMentor ($n=7$)}
      & $U$
      & $p_{\mathrm{raw}}$
      & $p_{\mathrm{Holm}}$
      & $r_{rb}$ \\
    \midrule
    Adaptive Diagnosis
      & $74.04\pm12.97$ & $80.36\pm9.83$
      & 30.0 & .189 & .378 & $-.34$ \\
    Instructional Responsiveness
      & $78.85\pm12.89$ & $64.29\pm13.36$
      & 72.0 & .030 & .090 & $.58$ \\
    Post-Revision Recovery
      & $75.00\pm14.43$ & $71.43\pm17.25$
      & 51.0 & .640 & .640 & $.12$ \\
    \bottomrule
  \end{tabular*}
\end{table*}

Five of the six AU interviewees described added learning that addressed
a difficulty they had encountered (T1, T2, T3, T5, T6).
T1 described more focused content:
\emph{``After replanning, the content was more focused and included
more concrete examples and questions. It was easier to understand
and more interesting.''} T2 described further explanations of concepts
and their differences. T5 valued the addition of unfamiliar technical
concepts. These accounts describe how targeted examples and added
concepts helped learners address specific difficulties and prepare
for later tasks.

Three AU interviewees described an easier retry or next step after
the added learning (T1, T2, T6). T2 recalled,
\emph{``After studying the added lesson, I answered again, and things
were slightly better.''} T1 also found later tasks easier to continue.
In the GenMentor group, G1 noted that explanations on the Replan page
were easy to overlook, while G2 found it difficult to distinguish
what had been reordered from what had been newly added.

Taken together, Study~3 provides complementary evidence about the
learning outcomes and experiences that arose as learners participated
in an evolving Capability Program. AU showed a larger Objective
Capability gain and stronger Cognitive Support, while the remaining
self-rating, experience, and revision comparisons were more tentative.
Interview accounts further illustrate how learners used feedback,
resource choices, and targeted revisions to revisit difficulties and
continue learning. These findings motivate the broader question of
how learner contributions can remain consequential as work passes
across coordinated roles.

\section{Discussion}
\label{sec:discussion}

AU offers a concrete way to organize educational work around a person whose capabilities, goals, and constraints can change through participation. The broader significance of this approach lies in sustaining human influence across agent roles and keeping the learning program aligned with the learner's current state and goals. We develop these implications from the system design and the complementary findings of our three studies.

\subsection{Human Participation as a Continuing Basis for Coordination}
\label{sec:discussion-commitments}

Participation in a multi-agent organization requires that what a person contributes remain connected to what different roles subsequently do. The formative interviews exposed the effort of repeatedly reconstructing learner context across tools and interactions. AU responds with two connected representations: the Human Node maintains the current account of the learner, and the Capability Program records the work organized around that account, so that a contribution made in one interaction remains relevant to later planning, mentoring, and assessment decisions.

Existing systems such as GenMentor establish the value of dynamic profiles and evolving paths~\cite{wang2025genmentor}. Our organizational perspective adds a criterion: meaningful participation depends on whether contributions remain consequential when work passes between roles. A constraint stated during planning, a difficulty encountered during practice, or a correction to an inferred background can each become a reason to reconsider future work. For HCI, AU offers a concrete design pattern for giving human input revisable effects that persist across roles within coordinated activity.

\subsection{Maintaining Learning Continuity as Learners Change}
\label{sec:discussion-actionable-adaptation}

Learning changes the conditions under which its own plan was formed: developing knowledge can reveal unrecognized prerequisites, and experience may alter what a learner considers useful or feasible. A program therefore needs to track the learner's current state while preserving the value of completed work---a tension the Capability Program makes explicit by linking goals, activities, assessment requirements, and revisions within a continuing trajectory.

The studies support both halves of this commitment. Study~1's clearest advantages appeared precisely in program organization and assessment design. Study~2 adds evidence about maintaining these relationships through change: AU received higher ratings for revision appropriateness, improvement, and coherence even though both systems received high evidence-alignment ratings, suggesting that useful adaptation depends on how a response to difficulty fits into subsequent learning rather than on reading evidence alone. This perspective makes the program's continuing suitability itself an object of evaluation: an unsuccessful attempt can prompt examination of prerequisite assumptions, activity demands, or available support alongside the learner's understanding, and revision can then preserve achievements while updating the conditions for further progress. The present studies demonstrate bounded program revision; sustained trajectories remain a direction for validation.

\subsection{Making Adaptive Organizations Accountable to Learners}
\label{sec:discussion-human-participation}

An organization that maintains an account of the learner also influences how progress is interpreted and which opportunities follow. Study~3 associated AU use with stronger cognitive support and larger objective assessment gains, while differences in self-efficacy and Metacognition gains were not statistically significant; interview accounts of fatigue and reconsidered understanding illustrate why performance and judgments of progress may diverge. Learners should therefore be able to see what particular evidence supports and how it informs subsequent work.

Correction and control~\cite{amershi2019guidelines} must accordingly hold across the whole organization, not only at the interface where a learner happens to be working---as when a background label overlooked a learner's prior training and shaped future opportunities. Corrections should affect the shared state used by subsequent roles, capability judgments should remain bounded by task and assistance conditions, and goals should stay open to reconsideration as priorities change. Shared representations make these relationships explicit enough to inspect and revise, and could support human educators and specialist agents working around a learner-owned program; our system-level results motivate this direction without isolating the effects of individual agents or establishing a general theory of human-participatory organization.

\subsection{From Personal Programs to Organizational Learning}
\label{sec:discussion-organizational-learning}

Although our studies evaluate only single-learner programs, the representations AU uses---Human Nodes, Capability Programs, and evidence records---also create conditions for learning \emph{across} learners. A Capability Program records not only what activities a learner completed but also which revisions were triggered, what evidence supported them, and whether those revisions led to subsequent success. Aggregated across learners pursuing related goals, these records could inform the planning role's initial path construction, where recurring prerequisite gaps suggest reordering, and the mentoring role's choice of explanations, where recurring misconceptions suggest addressing a difficulty before it arises. This organizational learning depends on the evidence discipline described above: capability claims must remain bounded by their conditions and learner corrections must remain legible, so that the organization improves through accountable use of participant evidence rather than opaque pattern-matching that reinforces earlier biases.% How to make such learning visible and contestable to the learners whose trajectories inform it is a question AU's design motivates but does not yet address.

% \subsection{Evidence Types and Their Organizational Consequences}
% \label{sec:discussion-evidence-types}

% The distinction between Learning-State Evidence and Capability Evidence is organizational rather than taxonomic: the two types justify different kinds of decisions. Learning-State Evidence---a question during a lesson, a request for hints, reported confidence---guides \emph{how} the organization supports the learner within a given activity. Capability Evidence is bounded by the task, the criteria, and the assistance conditions under which it was collected~\cite{messick1995validity}; it updates what the organization believes the learner has demonstrated and can justify changing \emph{which} activities follow.

% Conflating the two creates coordination failures across roles: a mentor acting on an inferred capability that no assessment has tested builds subsequent work on an unsupported assumption, while an assessor whose scoring absorbs unrecorded help overstates what the learner can do independently. AU's evidence model makes these boundaries explicit---each evidence item carries its source, conditions, and permitted updates---so that a role reading a capability claim also reads the scope within which it holds. For multi-agent learning systems, this suggests that evidence provenance and update rules are not implementation details but core organizational contracts.

\section{Limitations and Future Work}
\label{sec:limitations-future-work}

\paragraph{Empirical scope and validation.}
Our studies evaluated complete system configurations through selected synthetic cases and a short human study. Studies~1 and~2 used LLM-based evaluation under a fixed model configuration; Study~1 also examined correlations between expert and model ratings on a subset of cases. Evaluator models and scoring prompts may affect automated judgments~\cite{zheng2023judging}, motivating broader human review and comparisons across evaluators. Study~3 involved 32 learners aged 18--30, with recruitment focused on university students. Larger studies across age groups and racial, ethnic, and cultural communities could assess broader generalizability. Since systems differed in instructional formats and practice requirements, component studies and comparisons accounting for actual learning time could clarify the contributions of AU's organizational mechanisms, content formats, and practice opportunities.

\paragraph{Reliable evidence and meaningful participation.}
AU's decisions depend on the evidence carried through the Human Node. Errors in generated materials or assessment feedback may affect subsequent teaching and revision. AU records evidence sources and supports learner correction; how effectively these mechanisms resolve errors across roles requires further study. Independent reviews of instructional content and scoring could be combined with studies of concrete correction episodes: a learner challenges a diagnosis, updates a profile, or rejects a proposed change, and the organization responds. Following these episodes across mentoring and planning would show whether learner input has a lasting effect on subsequent work. Such studies could also examine how the organization handles conflicting evidence and when consultation with a human educator is useful.

\paragraph{Adaptation to different learners and changing goals.}
The human study used a researcher-assigned goal, while the revision benchmark examined prespecified low-performance conditions. The evaluated revisions primarily added support within an existing learning path. Everyday learning also involves changing direction, reducing workload, returning after interruptions, and deciding when further intervention is unnecessary. Future deployments could examine how learners and agents negotiate these choices as goals and available time change. Co-design with different learner groups, including learners with attention-deficit/hyperactivity disorder (ADHD), could inform adjustable task lengths, pacing, reminders, and feedback. Optional game elements, such as short missions and progress milestones, offer another direction for organizing practice. Their value should be examined through sustained learning and perceived control, with room for learners to change or disable them as their needs evolve.

\paragraph{Capability development beyond technical domains.}
Writing, the humanities, and workplace soft skills offer further settings in which to examine Capability Programs. The product-decision scenarios, written microprojects, and dialogue-based simulations in Study~3 provide initial examples of how AU can accommodate open-ended practice. They offer a starting point for dedicated studies of learning in these broader domains. Progress may become visible through successive drafts, stronger interpretations, or more effective responses to workplace situations. Assessment would need to capture these forms of development through domain-specific rubrics and expert or peer feedback. Differences in judgment could themselves become useful evidence for revision, particularly when several approaches are appropriate for the learner's goal.

\paragraph{Sustaining a personal learning organization.}
The longer-term vision of a ``university for each learner'' requires continuity across successive Capability Programs. Following learners across educational and career transitions would reveal when earlier evidence remains relevant and when reassessment is needed. Building on research in long-term agent memory~\cite{wu2024longmemeval,hu2025memoryagents,packer2023memgpt}, AU could consolidate experience, revisit outdated assumptions, and preserve the source and context of capability claims. Learners would need practical ways to inspect, correct, delete, and carry these records between programs. As goals change, the organization may also need to redistribute responsibilities among agents and involve human educators, making teacher handoffs and continuity of support useful subjects for study. Sustained use brings questions of generation cost and waiting time, which remain to be measured across the complete workflow. On-demand generation and reuse of suitable materials could help manage these costs. Longitudinal deployments would allow retention and transfer to be evaluated alongside the learner effort and resources required to sustain support.

\section{Conclusion}
\label{sec:conclusion}

We presented Agentic University as an instantiation of Human-Participatory Multi-Agent Organization. By connecting a correctable Human Node with a persistent Capability Program, AU allows learner contributions and assessed evidence to inform coordinated revisions to subsequent work. Three studies provide initial support for the integrated approach through evaluations of program quality, revision quality, and learner outcomes and experiences. The broader contribution is a design perspective in which human participation remains consequential across agent roles and the learning program remains revisable as the learner develops. Future work should examine whether this approach can sustain continuity, learner control, and reliable memory across open-ended, long-term learning trajectories.

\bibliographystyle{ACM-Reference-Format}
\bibliography{references}

\end{document}